\documentclass{article}

\usepackage{CJKutf8}
\usepackage{longcat/longcat_style}
\usepackage{adjustbox}
\usepackage[utf8]{inputenc} %
\usepackage[T1]{fontenc}    %
\usepackage{newunicodechar}
\usepackage{hyperref}       %
\usepackage{xcolor}
\usepackage[normalem]{ulem} %
\hypersetup{
    colorlinks=true,      %
    linkcolor=blue,      %
    urlcolor=blue,       %
    citecolor=blue,      %
    linkbordercolor=blue, %
    urlbordercolor=blue,
    citebordercolor=blue,
    pdfborderstyle={/S/U/W 1}, %
}
\usepackage{float}
\usepackage{url}            %
\usepackage{booktabs}       %
\usepackage{amsfonts}       %
\usepackage{nicefrac}       %
\usepackage{microtype}      %
\usepackage{lipsum}		%
\usepackage{graphicx}
\usepackage{xcolor}
\usepackage{colortbl}
\usepackage[most]{tcolorbox}
\usepackage{listings}
\usepackage{natbib}
\usepackage{doi}
\usepackage{amsmath}
\usepackage{amssymb} %
\usepackage{xspace}
\usepackage{enumitem}
\usepackage{multirow}
\usepackage{subcaption} 
\usepackage{makecell}
\usepackage{hyperref, cleveref}
\usepackage{pifont}
\usepackage[inkscapelatex=false]{svg}
\usepackage{caption}
\DeclareCaptionLabelSeparator{pipe}{ | }
\usepackage{tcolorbox}
\newtcolorbox{coloredquote}[1][]{
    colback=green!5!white,  % Background color
    colframe=green!70!black, % Border color
    boxrule=2pt,
    arc=7pt,
    left=6pt,
    right=6pt,
    top=4pt,
    bottom=4pt,
    title=#1
}

\setlist[itemize]{leftmargin=*}
\setlist[enumerate]{leftmargin=*}
\setlist[description]{leftmargin=*}
\usepackage{amsmath,amsfonts,bm}

\def\eqref#1{equation~\ref{#1}}
\def\1{\bm{1}}

\DeclareMathAlphabet{\mathsfit}{\encodingdefault}{\sfdefault}{m}{sl}
\SetMathAlphabet{\mathsfit}{bold}{\encodingdefault}{\sfdefault}{bx}{n}

\usepackage{amsmath}
\usepackage{amssymb}
\usepackage{booktabs}
\usepackage{graphicx}
\usepackage{placeins}
\usepackage{algorithm}
\usepackage{algorithmic}
\usepackage[table,dvipsnames]{xcolor}
\usepackage[most]{tcolorbox}
\usepackage{minted} % Local compilation requires -shell-escape.
\tcbuselibrary{listings, minted, breakable}
\usepackage{hyperref}
\usepackage{url}

\usepackage{amsthm}

\newtheorem{theorem}{Theorem}

\definecolor{figblue}{RGB}{46,102,156}
\definecolor{figorange}{RGB}{205,118,37}
\definecolor{figpurple}{RGB}{116,84,150}
\definecolor{figteal}{RGB}{36,132,125}

\newtcblisting{pycodebox}{
  listing engine=minted,
  minted language=python,
  minted style=vs,
  minted options={fontsize=\small, breaklines, tabsize=2},
  colback=blue!5!white,
  colframe=MidnightBlue!85!black,
  coltitle=white,
  colbacktitle=MidnightBlue!85!black,
  title=\textbf{Python Code},
  fonttitle=\bfseries,
  breakable,
  listing only
}

\newtcblisting{promptbox}[1]{
  listing engine=listings,
  listing only,
  listing options={
    basicstyle=\fontencoding{T1}\fontfamily{ptm}\selectfont\footnotesize,
    breaklines=true,
    breakatwhitespace=false,
    columns=fullflexible,
    keepspaces=true,
    showstringspaces=false
  },
  colback=SeaGreen!10!CornflowerBlue!10,
  colframe=RoyalPurple!55!Aquamarine!100!,
  title=\textbf{#1},
  fonttitle=\bfseries,
  breakable
}

\newtcolorbox{outputbox}{
  colback=gray!10!white,
  colframe=MidnightBlue!80!black,
  fonttitle=\bfseries,
  coltitle=white,
  colbacktitle=MidnightBlue!80!black,
  title=Output,
  left=4mm,
  right=4mm,
  top=2mm,
  bottom=2mm,
  boxrule=0.8pt,
  arc=3mm
}

\newcommand{\Mod}[1]{\textcolor{black}{#1}}

\title{MCTS Is Counting the Wrong Thing: Learning Exploration Counts for LLM Reasoning}

\author{%
    \textbf{%
    Xuan Zhang\textsuperscript{1,2,3}%
    \thanks{Equal contribution.}%
    \kern0.5em%
    \thanks{Work was done during the internship at Meituan.}\hspace{1.0em}%
    Zhijian Zhou\textsuperscript{1,2}\footnotemark[1]\hspace{1.0em}%
    Zicheng Hu\textsuperscript{1,3}\hspace{1.0em}%
    Yan Feng\textsuperscript{3}%
    } \\[2pt]
    \textbf{%
    Hongyan Hao\textsuperscript{3}\footnotemark[3]\hspace{1.0em}%
    Yuan Qi\textsuperscript{1,4}\footnotemark[3]\hspace{1.0em}%
    Chao Qu\textsuperscript{1,4}\thanks{Corresponding authors.}%
    } \\[3pt]
    \textsuperscript{1}Fudan University \quad
    \textsuperscript{2}Shanghai Innovation Institute \quad
    \textsuperscript{3}Meituan \\
    \textsuperscript{4}Shanghai Academy of Artificial Intelligence for Science \\
    \texttt{xuanzhang24@m.fudan.edu.cn \quad
    haohongyan02@meituan.com \quad
    \{qiyuan,quchao\}@fudan.edu.cn}
}

\renewcommand{\headeright}{\raisebox{-0.2\height}{\includegraphics[height=1.8em]{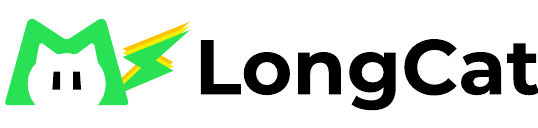}}}
\renewcommand{\shorttitle}{MCTS Is Counting the Wrong Thing: LPC-MCTS}

\begin{document}

\maketitle

% \begin{abstract}
% Effective exploration is central to Monte Carlo tree search (MCTS), yet
% standard UCT allocates search according to how often each tree branch has been
% visited. This mechanism becomes less informative in LLM reasoning, where states
% and actions are token sequences. Even small token changes create separate
% branches with independent counts, preventing continuations that follow similar
% reasoning patterns from sharing exploration evidence. Under a finite rollout
% budget, familiar strategies may therefore appear underexplored simply because
% they are worded differently. To address this limitation, we introduce Learned
% Pseudo-Count MCTS (LPC-MCTS), which replaces exact branch counts in the UCT
% exploration term with learned decision-level pseudo-counts derived from policy
% hidden representations. A lightweight Coin-Flipping Network (CFN), trained
% through random-label regression, estimates representation coverage at the token
% level and aggregates the scores into a node-level pseudo-count. This allows UCT
% to share evidence across representation-aligned branches and favor less-covered
% reasoning patterns. The learned pseudo-count can also prioritize less-covered trajectories for policy optimization.
% Experiments on agentic tasks, including tool-integrated mathematical reasoning
% and search, show that LPC-MCTS outperforms a strong MCTS baseline, demonstrating
% the effectiveness of learned pseudo-counts for guiding exploration.
% \end{abstract}

\begin{abstract}
Monte Carlo tree search (MCTS) has powered landmark successes in reinforcement learning, yet its gains in LLM reasoning can remain modest despite substantial search effort. We identify an overlooked bottleneck: \emph{what MCTS counts}. Standard UCT assigns independent visit counts to textual branches, allowing different formulations of the same reasoning strategy to repeatedly receive exploration bonuses. Our theoretical analysis shows that this \emph{count fragmentation} can impair exploration efficiency, as quantified by increased cumulative regret. To address this problem, we introduce Learned Pseudo-Count MCTS (LPC-MCTS), which changes only the exploration statistic in UCT. A lightweight Coin-Flipping Network head on the policy's hidden representations estimates representation coverage, enabling related branches to share exploration evidence. Across 5 mathematical-reasoning and 7 search benchmarks, LPC-MCTS consistently outperforms strong MCTS baselines. Holding each LPC-MCTS-trained policy fixed, replacing standard MCTS counts at inference improves average mean@16 accuracy across 5 mathematical benchmarks by 3.79 percentage points on Qwen2.5-7B and 2.71 points on Qwen3-4B. Applying LPC-MCTS during both training and inference yields gains of 4.96 and 4.37 points over standard MCTS, respectively. Together, our theory and experiments show that changing what MCTS counts can substantially improve its exploration efficiency and reasoning performance.
\end{abstract}

\section{Introduction}
\label{sec:introduction}

% Tree search has become an important way to improve LLM reasoning by
% exploring alternative continuations rather than committing to a single
% trajectory, both at inference time and during training
% \citep{yao2023tree,hao2023reasoning,zhou2023language,xie2024monte,
% wu2026deepsearch,ji2026tree}. Monte Carlo tree search (MCTS) provides a
% general framework for this process \citep{kocsis2006bandit,6145622}:
% rollout outcomes estimate branch values, while UCT computes its exploration
% term from empirical visit counts. These counts therefore play a central role
% in deciding how a limited search budget is allocated. In LLMs, however, each state is
% defined by a token sequence, so even small token differences produce distinct
% states with independent visit counts. Different
% branches can follow similar reasoning patterns, yet their visitation evidence
% remains separate, as illustrated by examples \textit{A} and \textit{B} in Figure \ref{fig:tree_local_vs_learned}. Under a finite budget, evidence is thus fragmented across
% many sparsely visited branches. When evidence cannot be shared, a limited search
% budget may be repeatedly assigned to familiar reasoning patterns expressed in
% slightly different language, leaving genuinely different strategies
% underexplored. 

Monte Carlo tree search (MCTS) is a powerful approach to sequential decision
making~\citep{kocsis2006bandit,6145622}. By combining search with learned
policies and value estimates, it helped enable landmark reinforcement learning
systems such as AlphaGo~\citep{silver2016mastering}. Its appeal extends
naturally to large language models (LLMs): reasoning requires choosing among
alternative plans, intermediate hypotheses, and tool calls, and search offers
a way to revisit these choices before committing to an answer.
A growing body of work has applied MCTS to LLM reasoning at inference time and
during training~\citep{hao2023reasoning,zhou2023language,qi2025mutual,
xie2024monte,zhang2024rest,wu2026deepsearch}. Meanwhile, methods such as Group
Relative Policy Optimization (GRPO) obtain strong reasoning policies from
groups of sampled responses without explicitly constructing search
trees~\citep{shao2024deepseekmath}. For MCTS to offer a compelling alternative,
its additional computation must translate into sufficiently better reasoning.

% To address this critical problem, count-based exploration offers a principled
% measure of familiarity: as visits accumulate, the exploration signal assigned
% to a state decreases. Exact counts, however, become uninformative in large or
% high-dimensional spaces where individual states rarely recur. Pseudo-count
% methods generalize this principle beyond exact state identity
% \citep{bellemare2016unifying,ostrovski2017count}, while related curiosity methods
% measure novelty through prediction error
% \citep{pathak2017curiosity,burda2018exploration}. Classical pseudo-count methods
% typically rely on density models that are costly to train and continually
% update. The Coin-Flipping Network (CFN) provides a lighter alternative by estimating pseudo-count
% through random-label regression, without requiring a normalized density model
% \citep{lobel2023flipping}. Recent applications to LLM preference optimization
% and reasoning further support the practicality of this approach
% \citep{bai2025online,zhang2026count}.

However, tree-search-based training can still yield only modest performance
gains over GRPO-style methods~\citep{wu2026deepsearch}. \citet{zhao2026reinforced} report that tree-based approaches such as
TreePO \citep{li2025treepobridginggappolicy} perform comparably to GRPO and its variants. The DeepSeek-R1
report also highlights the difficulty of iteratively improving model
performance through MCTS-based self-search~\citep{guo2025deepseek}.
Together, these findings indicate that using a tree structure alone does not
guarantee substantial gains in LLM reasoning. The effectiveness of exploration
within the tree therefore remains an important question. With this in mind,
our starting point is a component that is usually treated as routine
bookkeeping: the visit count. We identify a \emph{mismatch} between what MCTS
counts and what language reasoning needs to
explore. Our central finding is strikingly simple: \emph{changing only the
exploration count is enough to produce substantial gains.}

The issue arises from the meaning of a visit in a language search tree.
UCT allocates exploration according to branch values and visit counts,
giving larger exploration bonuses to less-visited branches
\citep{kocsis2006bandit}. When branches are identified by token sequences,
however, different formulations of the same strategy receive independent
counters. Repeatedly exploring one formulation does not increase the count of
another. The counts therefore record textual branch visits accurately while
failing to reflect how extensively a reasoning strategy has already been
explored. We refer to this mismatch as \emph{count fragmentation}. It allows
familiar strategies to repeatedly appear novel and compete for exploration
budget through different textual realizations. 

\begin{figure}[ht!]
    \begin{center}
    \includegraphics[width=0.8\linewidth]{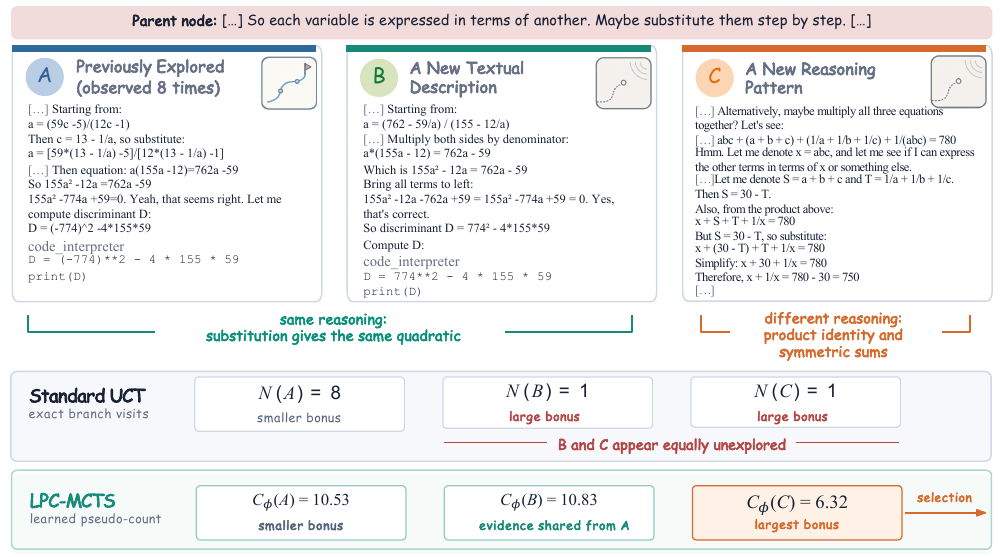}
    \end{center}
    \caption{Motivation for LPC-MCTS. The parent and branches \textit{A}-\textit{C}
    are taken from an actual model rollout tree, and the displayed $C_\phi$
    values are their learned pseudo-counts. Although branches \textit{A} and
    \textit{B} differ textually, they follow the same reasoning pattern and
    receive similar pseudo-counts, whereas \textit{C} follows a distinct
    pattern. Standard UCT instead treats \textit{B} and \textit{C} as equally
    unexplored because both have one exact visit. Additional rollout-tree
    examples are provided in Appendix~\ref{app:cfn_semantic_sensitivity}.}
    \label{fig:tree_local_vs_learned}
\end{figure}

Figure~\ref{fig:tree_local_vs_learned} illustrates this issue with an actual
rollout-tree example. Branches \textit{A} and \textit{B} follow the same
substitution-based derivation and reach the same quadratic, despite their
different textual forms, whereas branch \textit{C} instead uses a product
identity and symmetric sums. In standard UCT, although \textit{A} has already been explored
8 times, standard UCT treats the new branches \textit{B} and \textit{C}
as equally unexplored because both have one exact visit. LPC-MCTS instead
assigns similar pseudo-counts to \textit{A} and \textit{B} while separating
\textit{C}, allowing the search to share evidence across the two related
branches and prioritize the distinct reasoning pattern.

% Motivated by this connection, we introduce \emph{Learned Pseudo-Count MCTS}
% (LPC-MCTS), which replaces exact branch visits in the UCT exploration term with
% learned decision-level pseudo-counts that reflect familiarity across reasoning
% patterns. For each candidate decision, a CFN head scores its policy hidden
% representations, and the token-level scores are aggregated into a single
% pseudo-count for node selection. The pseudo-count is designed to
% assign weaker exploration signals to representation patterns judged more
% familiar, while preserving stronger signals for less-covered patterns.
% LPC-MCTS can therefore avoid spending rollouts on superficial
% rephrasings of familiar strategies and allocate more of the search budget to
% substantively new alternatives. Beyond tree selection, the learned pseudo-count can also prioritize less-covered trajectories for policy optimization.
% Since the estimator reuses the policy backbone and adds only a lightweight prediction head, it requires neither a separate density model nor another LLM backbone.

We address this problem with \emph{Learned Pseudo-Count MCTS} (LPC-MCTS),
which retains the UCT selection structure and replaces independent
branch-local visit counts with learned pseudo-counts. These counts estimate
representation coverage, allowing exploration evidence to be shared across
related branches. We obtain them using a Coin-Flipping Network
(CFN)~\citep{lobel2023flipping}, implemented as a lightweight auxiliary head
on top of the policy's existing hidden representations. \emph{The CFN head
adds almost no computational overhead to standard MCTS.} Evidence collected
from one branch can thereby increase the estimated familiarity of related
branches, reducing repeated exploration of already-covered reasoning
patterns and directing search toward less-explored alternatives.

% To our knowledge, LPC-MCTS is the first method to use learned pseudo-counts to
% guide MCTS exploration for LLM reasoning. Existing work improves LLM tree search
% through value estimation \citep{xie2024monte,wu2026deepsearch}, credit assignment
% \citep{ji2026tree,li2025treepobridginggappolicy}, pruning
% \citep{yang2026tooltree,qian2026paccfcalibratingirreversiblefrontier}, adaptive rollout allocation
% \citep{zhang2026maxs}. A closely related direction reduces redundant search by
% merging states according to embedding similarity \citep{wang2025don}.
% Meanwhile, recent count-based and curiosity-based methods primarily apply their
% exploration signals to policy optimization after responses have been sampled
% \citep{bai2025online,zhang2026count,dai2026cde}. LPC-MCTS instead places the
% learned count inside tree selection, where it determines which partial branch
% receives the next rollout. In the setting of
% \citet{lobel2023flipping}, the expected normalized squared CFN output is an
% inverse-coverage signal. We further show how exact count increments extend to
% graded evidence sharing between aligned hidden representations. As rollouts proceed,
% LPC-MCTS uses the state-visitation evidence accumulated by the CFN estimator to
% allocate relatively more exploration to branches assessed as less familiar.

We evaluate LPC-MCTS on two agentic settings: tool-integrated mathematical
reasoning and search. The results show that it improves task performance over
the strong MCTS baseline, indicating that representation-level coverage provides a more
informative basis for allocating search than independent counts over textual
branches. Our contributions are as follows:
\begin{itemize}
    \item We introduce LPC-MCTS, a novel tree-search method that replaces independent
    branch counts with learned decision-level pseudo-counts, enabling search
    evidence to generalize across related LLM reasoning patterns, thereby
    reducing redundant exploration.
    \item We theoretically characterize the exploration cost of count
    fragmentation through cumulative regret. 
    \item Our experiments on tool-integrated mathematical reasoning and search
    tasks demonstrate that LPC-MCTS improves task performance over strong MCTS
    baselines.
\end{itemize}

\section{Preliminaries}
\label{sec:preliminaries}

\subsection{MCTS and the role of counts}

MCTS repeatedly allocates computation through four stages
\citep{6145622}: (1) \emph{selection} follows promising edges from the
root; (2) \emph{expansion} adds an unexplored child; (3) \emph{simulation}
estimates its outcome; and (4) \emph{backpropagation} updates statistics along
the selected path. Repeating these stages concentrates simulations on strong branches while reserving some budget for alternatives whose value remains uncertain.

Let node $v$ represent state $s_v$, and let edge $(v,a,v_a)$ represent taking
action $a$ and reaching child $v_a$. In tool-augmented LLM
reasoning, the state is the interaction history; an action may be a reasoning
segment, a tool call, or a final answer, and a tool may append an external
observation. Let $Q(v,a)$ be the backed-up value of action $a$, $N(v)$ the
number of simulations through $v$, and $N(v_a)$ the number allocated to its
child branch. The UCT algorithm \citep{kocsis2006bandit} selects
\begin{equation}
    a^*=\arg\max_{a\in\mathcal A(v)}
    \left[Q(v,a)+c\sqrt{\frac{\log(N(v)+1)}{N(v_a)+\epsilon}}\right],
\end{equation}
where $c>0$ controls exploration and $\epsilon>0$ ensures numerical stability.
The first term favors high estimated value; the second is an optimistic bonus
that is larger for less-visited children.

\subsection{Coin-Flipping Network}

Exact visitation counts become uninformative in high-dimensional continuous
spaces because nearly every state representation may be unique. Classical
pseudo-count methods address this problem with learned density models
\citep{bellemare2016unifying,ostrovski2017count}, but training and continually
updating these models can be costly. A Coin-Flipping Network (CFN) provides a
lighter alternative by reducing count estimation to random-label regression
\citep{lobel2023flipping}. Whenever a representation
$h_i\in\mathbb{R}^{H}$ is observed, the CFN pairs it with a randomly sampled
Rademacher vector $z_i\in\{-1,+1\}^{d}$, whose coordinates independently take
either value with probability $1/2$. Importantly, repeated occurrences receive
different random targets. A predictor
$f_{\phi}:\mathbb{R}^{H}\rightarrow\mathbb{R}^{d}$ is trained by minimizing
the following objective:
\begin{equation}
    \mathcal{L}_{\mathrm{CFN}}(\phi)
    =\mathbb{E}_{(h_i,z_i)\sim\mathcal{D}_{\mathrm{CFN}}}
    \left[\left\|f_{\phi}(h_i)-z_i\right\|_2^2\right],
\end{equation}
where $\mathcal{D}_{\mathrm{CFN}}$ contains the observed
representation-target pairs. If the representation $h$ occurs
$n$ times, mean-squared loss minimization predicts the mean of its targets,
$f^*(h)=\frac{1}{n}\sum_{i=1}^{n}z_i$. Since each target coordinate is independent,
zero-mean, and unit-variance,
\begin{equation}
    \mathbb{E}\left[\frac{1}{d}\left\|f^*(h)\right\|_2^2\right]=\frac{1}{n}.
    \label{eq:cfn_count_identity}
\end{equation}
Thus, $\frac{1}{d}\|f_\phi(h)\|_2^2$ estimates an \emph{inverse count}: it
shrinks as evidence accumulates. The $d$ coordinates estimate the same inverse
count independently in this idealized solution, so averaging them reduces
estimator variance. Taking its square root gives the CFN count-based bonus
$b_\phi(h)=\sqrt{\frac{1}{d}\|f_\phi(h)\|_2^2}$, which scales as $\frac{1}{\sqrt{n}}$
under repeated observations.

\section{Methodology}
\label{sec:method}

\begin{figure}[ht!]
    \begin{center}
    \includegraphics[width=0.9\linewidth]{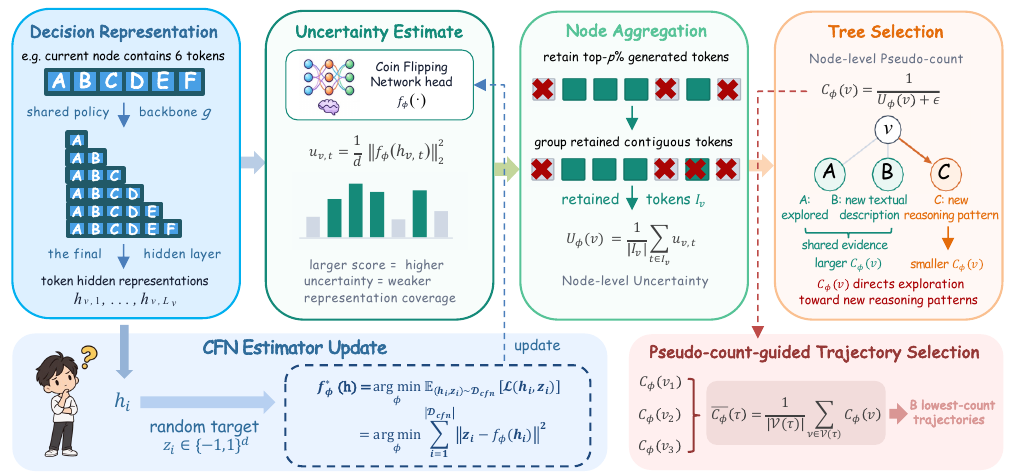}
    \end{center}
    \caption{Overview of LPC-MCTS, including CFN-based pseudo-count estimation, estimator training, tree selection, and pseudo-count-guided trajectory selection.}
    \label{fig:lpc_mcts_overview}
\end{figure}

LPC-MCTS replaces the visit count with a learned pseudo-count
that measures familiarity across model decisions. As shown in
Figure~\ref{fig:lpc_mcts_overview}, a CFN estimator computes this count from
hidden representations, and the count directly determines the
exploration bonus used for MCTS selection.

\vspace{-0.5em}

\subsection{Learned decision-level pseudo-counts}
\label{sec:learned_evidence}
\label{sec:token_count}

\paragraph{CFN estimation and training.}
We attach a lightweight linear CFN head to the policy's hidden
representations. For $h\in\mathbb{R}^{H}$, we compute
\vspace{-0.3em}
\begin{equation}
    f_\phi(h)=W_\phi^\top\operatorname{sg}(h),
    \qquad
    u_\phi(h)=\frac{1}{d}\|f_\phi(h)\|_2^2,
    \label{eq:cfn_head_score}
\end{equation}
where $W_\phi\in\mathbb{R}^{H\times d}$ is trainable and
$\operatorname{sg}(\cdot)$ prevents the CFN loss from updating the policy
backbone. Following \citet{lobel2023flipping}, we use $u_\phi(h)$ as a
token-level inverse-count signal. During search, we collect representations
at model-generated reasoning and tool-call positions, excluding prompt and
tool-response tokens. Each observed occurrence $h_i$ is paired with a fresh,
independently sampled Rademacher target $z_i\in\{-1,+1\}^{d}$. We train the
head on minibatches $\mathcal B_{\mathrm{CFN}}$ of the accumulated pairs using
\vspace{-0.2em}
\begin{equation}
    \mathcal L_{\mathrm{CFN}}(\phi)
    =\frac{1}{|\mathcal B_{\mathrm{CFN}}|}
     \sum_{(h_i,z_i)\in\mathcal B_{\mathrm{CFN}}}
     \left\|f_\phi(h_i)-z_i\right\|_2^2.
    \label{eq:cfn_estimator}
\end{equation}
\vspace{-1.2em}

\paragraph{From token scores to a node count.}
For a search node $v$, let
$x_v=(x_{v,1},\ldots,x_{v,L_v})$ denote its token sequence. We encode this
sequence independently with the policy backbone $g$ and score each position
with the trained CFN head $f_\phi$:
\vspace{-0.5em}
\begin{equation}
    h_{v,t}=g(x_{v,\leq t}),
    \qquad
    u_{v,t}=\frac{1}{d}\|f_\phi(h_{v,t})\|_2^2.
\end{equation}
\vspace{-1.0em}

By the definition above, $u_{v,t}$ is the representation-level uncertainty at
position $t$. A larger score means that the CFN has accumulated less evidence
for the represented state, making that position more important for exploration.
Among the model-generated tokens in node $v$, we retain the top-$p\%$ with the
largest $u_{v,t}$, so that the node
statistic emphasizes the exploration signal from positions with the relatively higher
uncertainty. Since isolated high scores may be noisy, we
partition the retained positions into contiguous runs and discard short runs
that do not form coherent local segments.
Let $I_v$ be the union of the remaining runs. We then define
\vspace{-0.2em}
\begin{equation}
    U_{\phi}(v)=\frac{1}{|I_v|}\sum_{t\in I_v}u_{v,t},
    \qquad
    C_{\phi}(v)=\frac{1}{U_{\phi}(v)+\epsilon}.
    \label{eq:node_pseudocount}
\end{equation}
$U_\phi(v)$ summarizes uncertainty over coherent high-score regions. 
$C_\phi(v)$ is the reciprocal of $U_\phi(v)+\epsilon$. We use it as a learned
pseudo-count: a larger uncertainty score produces a smaller $C_\phi(v)$ and
thus, all else being equal, a larger exploration bonus in
Eq.~\ref{eq:learned_uct}.

\subsection{Pseudo-count-guided tree search}
\label{sec:tree_policy}

\paragraph{Constructing tree nodes.}
During expansion, we grow the search tree together with the generated
trajectory and use token entropy as the signal for creating new nodes. After
each tool response, the policy generates a probe window of $w$ tokens. Let
$\overline H_{\mathrm{win}}$ and $\overline H_{\mathrm{ctx}}$ denote the mean
token entropy over the probe window and the preceding context, respectively.
We create a new node when
$\overline H_{\mathrm{win}}>\alpha\overline H_{\mathrm{ctx}}$;
otherwise, generation continues within the current node. Here, $\alpha>1$ controls
the frequency of node branching, with a larger $\alpha$ leading to less
frequent branching. 

\paragraph{Shared rollout and tree values.}
During expansion, the behavior policy $\pi_{\theta_{\mathrm{old}}}$ generates
candidate action segments. Repeated selection and expansion yield terminal
trajectories. A verifier or reward model assigns each completed trajectory
$\tau$ an outcome reward $R(\tau)$, and these outcomes are used to maintain an
action value $Q(v,a)$ for each expanded edge $(v,a)$.

\paragraph{UCT-style acquisition rule.}
At a node $v$ with expanded children, let
$\mathcal{A}_{\mathrm{active}}(v)$ denote their actions and let $v_a$ be the
child reached by action $a$. We place the parent
normalizer on the same learned scale as the child pseudo-counts by setting
$M_\phi(v)=\sum_{a\in\mathcal{A}_{\mathrm{active}}(v)}C_\phi(v_a)$.
LPC-MCTS selects
\vspace{-0.1em}
\begin{equation}
    a^* = \arg\max_{a\in\mathcal{A}_{\mathrm{active}}(v)}
    \left[
        Q(v,a) + c_{\mathrm{count}}
        \sqrt{
            \frac{\log(M_{\phi}(v)+1)}
                 {C_{\phi}(v_a)}
        }
    \right].
    \label{eq:learned_uct}
\end{equation}
By construction, at a fixed selection step, a larger node score
$U_\phi(v_a)$ gives a smaller $C_\phi(v_a)$ and hence a larger exploration
bonus.

\paragraph{Policy optimization from searched trajectories.}
Following the node-value formulation of DeepSearch
\citep{wu2026deepsearch} and group-relative reward normalization
\citep{shao2024deepseekmath,chu2026gpg}, we compute an action-level advantage
from the corresponding tree value.
Let $\mathcal{L}_x$ denote the terminal trajectories for prompt $x$:
\vspace{-0.2em}
\begin{equation}
    b_x=\frac{1}{|\mathcal{L}_x|}\sum_{\tau\in\mathcal{L}_x}R(\tau),
    \qquad
    A(v,a)=Q(v,a)-b_x.
    \label{eq:tree_advantage}
\end{equation}
Here $Q(v,a)$ denotes the tree value associated with edge $(v,a)$, and $b_x$ is the prompt-level mean terminal outcome used as the baseline. All
generated tokens in an action segment $a=(y_1,\ldots,y_{L_a})$ share this
action-level advantage.

We optimize these action segments using the clipped surrogate objective of PPO
\citep{schulman2017proximal}. For token $t$, the importance ratio is
$
    r_{\theta,t}(v,a)
    =\frac{\pi_\theta(y_t\mid p_v,y_{<t})}
           {\pi_{\theta_{\mathrm{old}}}(y_t\mid p_v,y_{<t})},
$
where $p_v$ is the complete prefix at node $v$. Following DAPO
\citep{yu2025dapo}, we use separate lower and upper clipping ranges and average
the surrogate over all trainable tokens:
\begin{equation}
\scalebox{0.95}{$\displaystyle
\begin{aligned}[b]
    \mathcal J_{\mathrm{policy}}(\theta)
    =\frac{1}{\sum_{(v,a)\in\mathcal D_{\mathrm{tree}}}L_a}
     \sum_{(v,a)\in\mathcal D_{\mathrm{tree}}}\sum_{t=1}^{L_a}
    \min\Big(&r_{\theta,t}(v,a)\,A(v,a),\\
             &\operatorname{clip}\!\left(
                 r_{\theta,t}(v,a),
                 1-\epsilon_{\mathrm{low}},
                 1+\epsilon_{\mathrm{high}}
              \right)A(v,a)\Big),
\end{aligned}
$}
\label{eq:policy_objective}
\end{equation}
We additionally use dual clipping for negative advantages to limit extreme
updates from large importance ratios, as detailed in Appendix~\ref{app:policy_optimization}.
LPC-MCTS influences policy learning through
exploration: the pseudo-count changes which branches are searched and hence
which action segments enter $\mathcal D_{\mathrm{tree}}$, while empirical
visits remain in use for value averaging.

\subsection{Pseudo-count-guided trajectory selection}
\label{sec:terminal_selection}

Let $\mathcal C_x=\{\tau_1,\ldots,\tau_K\}$ be the candidate pool of completed
trajectories for prompt $x$, and let $\mathcal V(\tau)$ denote the set of
nodes on trajectory $\tau$. For trajectory comparison, node scores within a
candidate pool are evaluated against a shared reference state of $f_\phi$. We
define the pseudo-count of a trajectory as the mean of its node-level
pseudo-counts:
\begin{equation}
    \overline C_\phi(\tau)
    =\frac{1}{|\mathcal V(\tau)|}
      \sum_{v\in\mathcal V(\tau)} C_\phi(v), \qquad
    \mathcal B_x
    =\underset{\substack{\mathcal B\subseteq\mathcal C_x\\|\mathcal B|=B}}
      {\arg\min}
      \sum_{\tau\in\mathcal B}\overline C_\phi(\tau).
\label{eq:trajectory_selection}
\end{equation}
We sort $\mathcal C_x$ by $\overline C_\phi(\tau)$ in ascending order and pass
the first $B$ trajectories, denoted by $\mathcal B_x$, to policy optimization.
This rule prioritizes trajectories that pass through
underexplored decisions. It thereby extends pseudo-count-guided exploration
from tree expansion to training-batch construction.

\section{How Count Fragmentation Impairs Exploration}
\label{sec:count_fragmentation_regret}

We next analyze how count fragmentation affects exploration efficiency.
We quantify this effect through \emph{cumulative regret}, a standard
criterion in online learning that measures the reward lost relative to
always selecting an optimal action~\citep{auer2002finite,lattimore2019bandit}.
We begin with the simplest setting, a one-step decision problem modeled
as a stochastic bandit, and show that count fragmentation already incurs
an additional exploration cost.

Consider a fixed finite bandit with $K\ge2$ groups of actions.
Within each group, actions are distinct choices but yield rewards from
the same distribution. Let $\mathcal C_c$ denote group $c$, containing
$m_c=|\mathcal C_c|$ actions. Rewards lie in $[0,1]$, are independent
across pulls, and have mean $\mu_c$ for every action in group $c$.
Assume group $1$ contains the unique optimal action, so $m_1=1$ and
$\mu_1>\mu_c$ for all $c\ge2$. Let $\Delta_c=\mu_1-\mu_c$ denote the
reward gap of a suboptimal group. For actions $A_1,\ldots,A_T$ selected
over $T$ rounds, the expected cumulative regret is
$R(T)
    =\mathbb E\!\left[
        \sum_{t=1}^{T}\bigl(\mu_1-\mu_{c(A_t)}\bigr)
      \right]
    =\sum_{c=2}^{K}\Delta_c\,\mathbb E[N_c(T)],
$
where $c(a)$ identifies the group containing action $a$, and $N_c(T)$
counts all selections of actions in group $c$. Regret therefore measures
the sampling budget spent on suboptimal actions, weighted by their
reward gaps.

\paragraph{Independent versus shared counts.}
Let $N_a(t)$ and $\widehat\mu_a(t)$ denote the visit count and empirical
mean of action $a$ after $t$ rounds, and define the class count as
$N_c(t)=\sum_{a\in\mathcal C_c}N_a(t)$.
We compare two UCB rules that differ only in their exploration counts.
Standard UCB1 uses
$I_{\mathrm{ind}}(a,t)=\widehat\mu_a(t-1)
+\sqrt{2\log t/N_a(t-1)}$,
whereas the shared-count variant uses
$I_{\mathrm{share}}(a,t)=\widehat\mu_a(t-1)
+\sqrt{2\log t/N_{c(a)}(t-1)}$.
Both sample every action once before selecting actions with the largest
index. Each action retains its own empirical mean, computed only from
its observed rewards; only the exploration counts are shared.

\begin{theorem}[Regret penalty of count fragmentation]
\label{theorem:count_fragmentation_regret}
For the bandit and selection rules defined above, as $T\to\infty$,
\begin{equation}
    R_{\mathrm{ind}}(T)
    =2\log T\sum_{c=2}^{K}\frac{m_c}{\Delta_c}+o(\log T),
    \qquad
    R_{\mathrm{share}}(T)
    =2\log T\sum_{c=2}^{K}\frac{1}{\Delta_c}+o(\log T).
    \label{eq:cf_asymptotics}
\end{equation}
In particular, if every suboptimal class contains $m$ equivalent actions,
then $R_{\mathrm{ind}}(T)/R_{\mathrm{share}}(T)\to m$.
\end{theorem}

Independent counts therefore repeat the same logarithmic exploration
cost across equivalent actions, increasing regret even when rewards
are deterministic. The proof and a finite-time bound are provided in
Appendix~\ref{app:count_fragmentation_regret}.

\paragraph{Extension to tree search.}
We extend our analysis to recursive UCT with terminal-reward backup
and obtain a similar regret separation. In the tree setting,
the leading exploration costs accumulate along the optimal path,
with independent counts paying separately for each equivalent action.
When every suboptimal class on this path contains $m$ equivalent actions,
independent-count UCT incurs $m$ times the leading cumulative regret
of shared-count UCT, while both retain separate action-value estimates.
The formal statement (Theorem \ref{prop:tree_count_fragmentation}), assumptions, and proof are provided in
Appendix~\ref{app:tree_count_fragmentation}.

\vspace{-0.2em}
\section{Related Work}
\label{sec:related_work}

\paragraph{Tree-structured search for LLM agents.}
Tree of Thoughts and RAP formulate language reasoning as explicit search over
intermediate states \citep{yao2023tree,hao2023reasoning}, while ReAct and LATS
extend this view to agents that interleave reasoning, actions, and environmental
observations \citep{yao2023react,zhou2023language}.
Subsequent work augments tree search with tool-aware feedback, reflective
simulation, safe pruning, adaptive lookahead, and multi-agent reasoning
\citep{yang2026tooltree,zhang2026spiral,qian2026paccfcalibratingirreversiblefrontier,
zhang2026maxs,li2026llm}. Recent training-time methods use the tree structure
to obtain backed-up node values or fine-grained relative advantages, improving
rollout efficiency and credit assignment
\citep{wu2026deepsearch,ji2026tree,li2025treepobridginggappolicy}. LPC-MCTS addresses a
complementary question: how exploration evidence should be represented during
tree selection. Some methods reduce redundant exploration through
embedding-based clustering and merging of semantically similar search states. For instance, 
FETCH uses a separate embedding model to form
these clusters~\citep{wang2025don}. Compared with this class of explicit
clustering methods, LPC-MCTS requires neither an additional embedding model nor
online clustering. Its lightweight CFN head operates on the policy's existing hidden
representations and is updated with states collected during rollouts. The
pseudo-count therefore reflects how familiar a state is under the current
policy after accumulated rollout visitation and enters tree selection directly.

\paragraph{Count-based exploration beyond tabular spaces.}
UCT treats the visit count as the evidence supporting a branch
\citep{kocsis2006bandit}. Exact counts are effective when search repeatedly
reaches the same discrete states, but language trees contain many low-visit
nodes whose textual forms differ even when their reasoning content is similar.
Generalized pseudo-counts address this problem in large observation spaces by
replacing exact identity with learned density or representation-level
recurrence \citep{bellemare2016unifying,ostrovski2017count}. Coin-Flipping
Networks (CFNs) estimate such recurrence through random-label regression,
avoiding an explicit normalized density model \citep{lobel2023flipping}.
Recent work has brought these exploration signals to LLM training. COPO uses a
CFN pseudo-count over prompt--response pairs for optimistic preference
alignment \citep{bai2025online}, while MERCI applies token-level CFN scores as
intrinsic rewards for diverse reasoning trajectories \citep{zhang2026count}.
CDE instead combines actor surprise with critic disagreement
\citep{dai2026cde}. These methods shape policy learning after responses are
sampled, whereas LPC-MCTS uses
the learned pseudo-counts directly during tree selection.

\vspace{-0.2em}
\section{Experiments}
\label{sec:experiments}
To evaluate whether learned pseudo-counts improve exploration and final task
performance, we study LPC-MCTS in two agentic settings: tool-integrated
mathematical reasoning and search.

\vspace{-0.2em}
\subsection{Experimental Setup}

\paragraph{Tool-integrated mathematical reasoning.}
We use Qwen2.5-7B \citep{qwen2025qwen25technicalreport} and Qwen3-4B
\citep{yang2025qwen3} as policy backbones. Both models are initialized with the
same ReTool SFT dataset \citep{feng2026retool}. For RL training, we use
DAPO-Math-17K \citep{yu2025dapo}. We evaluate on AIME
2024/2025/2026 \citep{aime}, as well as HMMT February 2025 and BRUMO 2025
from MathArena \citep{dekoninck2026matharena}. For each problem, we sample 16 responses with temperature $T=0.6$ and top-$p=0.95$,
and report their mean accuracy (mean@16).

\paragraph{Search.}
We use Qwen2.5-7B-Instruct and Llama-3.2-3B-Instruct
\citep{grattafiori2024llama3herdmodels} as policy backbones. Both models are
initialized by SFT on the same 2,000 examples sampled from the dataset
constructed by \citet{li2025chainofagentsendtoendagentfoundation}. Following
\citet{jin2025searchr}, our RL training set consists of the training splits of
NQ \citep{kwiatkowski-etal-2019-natural} and HotpotQA
\citep{yang2018hotpotqadatasetdiverseexplainable}. We evaluate our training
paradigm on 7 search-augmented QA tasks, including single-hop QA datasets (NQ,
TriviaQA \citep{joshi2017triviaqalargescaledistantly}, and PopQA
\citep{mallen2023trustlanguagemodelsinvestigating}) and multi-hop QA datasets
(HotpotQA, 2Wiki \citep{ho2020constructingmultihopqadataset}, MuSiQue
\citep{trivedi2021musique}, and Bamboogle
\citep{press2023measuringnarrowingcompositionalitygap}). We use offline
retrieval over the 2018 Wikipedia corpus \citep{karpukhin-etal-2020-dense}
using E5-base-v2 embeddings
\citep{wang2024textembeddingsweaklysupervisedcontrastive}, and return the top-3 documents for each query.
We evaluate each question with greedy sampling and MCTS.
For MCTS, we set the temperature $T=0.6$ and top-$p=0.95$, and report the mean accuracy over 5 responses per question (mean@5).

\paragraph{Baselines and configurations.}
We use DAPO as our policy-optimization baseline and additionally include \Mod{three}
exploration-oriented methods. \citet{cheng2025reasoningexplorationentropyperspective}
encourage exploration through entropy-based advantage shaping; this baseline
is denoted as DAPO + Entropy. \citet{gao2026navigateunknownenhancingllm} use
random network distillation (RND)-based intrinsic rewards for exploration;
this baseline is denoted as DAPO + IMAGINE.
\Mod{\citet{zhang2026count} uses CFN-based intrinsic rewards to encourage exploration of unfamiliar reasoning trajectories; this baseline is denoted as DAPO + MERCI.}
We compare these baselines with standard visit-count MCTS and
LPC-MCTS, which jointly trains the policy and CFN head.
During tree search, standard MCTS uses discrete visit counts, whereas LPC-MCTS
replaces them with CFN-estimated pseudo-counts. \Mod{At inference time, the CFN head is frozen, so its pseudo-counts reflect coverage learned during training and do not update with test-time visits.}
For MCTS-based inference, motivated by the use of
self-consistency-derived terminal rewards for MCTS backup in
rStar~\citep{qi2025mutual}, we assign terminal rewards through answer consistency
without using reference answers. After adding each terminal node, we group all
terminal nodes by their final answers, assign $+1$ to nodes in the
current majority group and $-1$ to the others, and recompute the backed-up
$Q$-values throughout the tree.
Details are provided in
Appendix~\ref{app:training_configuration}.

\begin{table*}[t]
  \caption{Mathematical reasoning results under different training and
  test-time inference settings. \Mod{SFT denotes the model before RL training.} The \textbf{bold text} represents the best, while \underline{underlined} represents the second best.}
  \label{tab:math_main}
  \begin{center}
  \resizebox{\textwidth}{!}{%
  \begin{tabular}{llcccccc}
    \toprule
    Training & Inference & AIME 24 & AIME 25 & AIME 26 & HMMT25 & BRUMO25 & Avg. \\
    \midrule
    \multicolumn{8}{l}{\textit{Qwen2.5-7B}} \\
    \addlinespace
    \Mod{SFT} & \Mod{Direct sampling} & \Mod{9.58} & \Mod{8.54} & \Mod{8.13} & \Mod{4.17} & \Mod{19.58} & \Mod{10.00} \\
    \Mod{SFT} & \Mod{MCTS} & \Mod{10.42} & \Mod{12.08} & \Mod{8.54} & \Mod{10.00} & \Mod{21.04} & \Mod{12.42} \\
    \midrule
    DAPO             & Direct sampling & 24.58 & 23.33 & 20.21 & 9.38 & 30.63 & 21.63 \\
    DAPO             & MCTS            & 29.79 & 27.71 & 23.33 & 16.25 & 40.63 & 27.54 \\
    DAPO + Entropy   & Direct sampling & 27.08 & 22.92 & 23.96 & 12.29 & 33.33 & 23.92 \\
    DAPO + Entropy   & MCTS            & 30.42 & 27.50 & \underline{26.25} & 14.17 & 42.08 & 28.08 \\
    DAPO + IMAGINE   & Direct sampling & 24.38 & 21.46 & 20.21 & 10.83 & 32.08 & 21.79 \\
    DAPO + IMAGINE   & MCTS            & 28.33 & 26.46 & 23.96 & 16.04 & 41.88 & 27.33 \\
    \Mod{DAPO + MERCI} & \Mod{Direct sampling} & \Mod{25.42} & \Mod{23.54} & \Mod{22.29} & \Mod{12.92} & \Mod{33.33} & \Mod{23.50} \\
    \Mod{DAPO + MERCI} & \Mod{MCTS} & \Mod{\underline{32.29}} & \Mod{28.75} & \Mod{25.21} & \Mod{14.38} & \Mod{40.00} & \Mod{28.13} \\
    MCTS             & Direct sampling & 25.21 & 23.75 & 21.04 & 11.25 & 30.42 & 22.33 \\
    MCTS             & MCTS            & 28.33 & 27.50 & 26.04 & 15.21 & 42.08 & 27.83 \\
    \midrule
    LPC-MCTS         & Direct sampling & 26.67 & 24.38 & 22.08 & 13.33 & 34.79 & 24.25 \\
    LPC-MCTS         & MCTS            & 30.83 & \underline{29.17} & 25.42 & \underline{17.08} & \underline{42.50} & \underline{29.00} \\
    \rowcolor{CornflowerBlue!12}
    LPC-MCTS & LPC-MCTS & \textbf{37.08} & \textbf{30.63} & \textbf{27.92} & \textbf{23.33} & \textbf{45.00} & \textbf{32.79} \\
    \midrule[\heavyrulewidth]
    \multicolumn{8}{l}{\textit{Qwen3-4B}} \\
    \addlinespace
    \Mod{SFT} & \Mod{Direct sampling} & \Mod{40.42} & \Mod{28.33} & \Mod{25.42} & \Mod{20.63} & \Mod{37.50} & \Mod{30.46} \\
    \Mod{SFT} & \Mod{MCTS} & \Mod{36.46} & \Mod{25.00} & \Mod{18.96} & \Mod{16.88} & \Mod{39.58} & \Mod{27.38} \\
    \midrule
    DAPO             & Direct sampling & 59.58 & 51.46 & 54.17 & 36.25 & 58.33 & 51.96 \\
    DAPO             & MCTS            & 71.88 & 57.50 & 59.38 & 47.92 & 61.67 & 59.67 \\
    DAPO + Entropy   & Direct sampling & 60.63 & 53.33 & 55.83 & 35.63 & 61.46 & 53.38 \\
    DAPO + Entropy   & MCTS            & 69.17 & 56.04 & 64.17 & 43.96 & \textbf{68.33} & 60.33 \\
    DAPO + IMAGINE   & Direct sampling & 62.92 & 49.38 & 57.29 & 37.08 & 58.75 & 53.08 \\
    DAPO + IMAGINE   & MCTS            & 69.58 & 55.42 & 61.46 & 49.38 & 65.42 & 60.25 \\
    \Mod{DAPO + MERCI} & \Mod{Direct sampling} & \Mod{63.13} & \Mod{52.29} & \Mod{58.33} & \Mod{35.63} & \Mod{59.79} & \Mod{53.83} \\
    \Mod{DAPO + MERCI} & \Mod{MCTS} & \Mod{70.83} & \Mod{57.92} & \Mod{60.83} & \Mod{49.17} & \Mod{61.46} & \Mod{60.04} \\
    MCTS             & Direct sampling & 60.83 & 52.29 & 57.50 & 35.83 & 60.00 & 53.29 \\
    MCTS             & MCTS            & \underline{73.75} & 56.46 & 65.00 & 50.21 & 65.21 & 62.13 \\
    \midrule
    LPC-MCTS         & Direct sampling & 62.71 & 53.13 & 58.13 & 36.67 & 59.38 & 54.00 \\
    LPC-MCTS         & MCTS            & 71.25 & \underline{59.17} & \textbf{74.38} & \underline{50.63} & 63.54 & \underline{63.79} \\
    \rowcolor{CornflowerBlue!12}
    LPC-MCTS & LPC-MCTS & \textbf{78.13} & \textbf{63.54} & \underline{72.92} & \textbf{51.88} & \underline{66.04} & \textbf{66.50} \\
    \bottomrule
  \end{tabular}%
  }
  \end{center}
\end{table*}

\subsection{Analysis of the Learned CFN Head}
\label{sec:cfn_analysis}

% We examine the learned CFN head from two complementary perspectives.

\paragraph{Token-level uncertainty visualization.}
Within each selected response, we visualize how the CFN score $u_{v,t}$ of every
eligible generated token changes across the sequence.
The visualizations and corresponding analysis are provided in
Appendix~\ref{app:cfn_token_visualization}.
Across these examples, high scores cluster on novel reasoning patterns, such as introducing a new solution strategy, applying a theorem, correcting an earlier step, reformulating a query, or integrating evidence into an answer.
In contrast, common connecting text and simple calculations or routine queries receive lower scores.
The high values also extend across groups of nearby tokens rather than appearing as isolated spikes.
Together, these observations suggest that the learned CFN head assigns higher uncertainty to novel high-level reasoning patterns in both mathematical reasoning and search, rather than isolated token differences.
This behavior directly supports our learned pseudo-count: Eq.~\ref{eq:node_pseudocount} maps less-familiar reasoning spans with larger $U_\phi(v)$ to a smaller $C_\phi(v)$ and thus a larger exploration bonus in Eq.~\ref{eq:learned_uct}, while familiar patterns receive larger learned counts and less exploration.

\paragraph{Reasoning-pattern sensitivity.}
To test whether the learned pseudo-count captures reasoning-pattern similarity
during search, we sample some sibling groups from actual model rollout trees.
In each group, two textually different children follow the same or a closely
related reasoning pattern, while a third child takes a distinct approach. The
two pattern-aligned children receive closely matched pseudo-counts, whereas the
distinct child is separated from both by a markedly larger gap. This separation
indicates that the estimator responds to changes in reasoning pattern rather
than surface-level wording alone. In these cases, the distinct child also
receives a lower pseudo-count and therefore a larger exploration bonus in
Eq.~\ref{eq:learned_uct}. The complete cases are analyzed in
Appendix~\ref{app:cfn_semantic_sensitivity}. These results qualitatively support
the motivation in Figure~\ref{fig:tree_local_vs_learned}: discouraging redundant
expansion of familiar patterns while preserving exploration for distinct
strategies.

\paragraph{\Mod{Quantitative analysis of evidence sharing.}}
\Mod{We quantify semantic alignment and count transfer in mathematical reasoning using Qwen3-4B and Qwen2.5-7B in Appendix~\ref{app:cfn_mechanism_validation}.
Pseudo-count closeness correlates positively with embedding similarity ($\rho=0.248$--$0.358$), and nearest-neighbor agreement reaches 24.1--29.2\%, above the 14.3\% uniform-random baseline.
These correlations suggest that learned pseudo-counts provide an exploration signal that is related to semantic similarity but does not fully match it.
In the transfer evaluation, 16 CFN updates on one branch increase the pseudo-counts of held-out same-pattern siblings by 5.5\%, while those of distinct-pattern siblings decrease by 3.2\%.
Together, these results support the use of learned pseudo-counts to share exploration evidence across related reasoning branches.}

\subsection{Main Results}

\begin{table*}[t]
  \caption{Search results under different training and test-time inference
  strategies. \Mod{SFT denotes the model before RL training.} We evaluate greedy sampling with temperature $T=0$ and MCTS with temperature $T=0.6$.}
  \label{tab:search_main}
  \begin{center}
  \resizebox{\textwidth}{!}{%
  \begin{tabular}{llcccccccc}
    \toprule
    Training & Inference & NQ & TriviaQA
      & PopQA & HotpotQA & 2Wiki
      & MuSiQue & Bamboogle & Avg. \\
    \midrule
    \multicolumn{10}{l}{\textit{Qwen2.5-7B-Instruct}} \\
    \addlinespace
    \Mod{SFT} & \Mod{Greedy sampling} & \Mod{36.80} & \Mod{54.40} & \Mod{25.80} & \Mod{43.20} & \Mod{38.60} & \Mod{29.20} & \Mod{46.40} & \Mod{39.20} \\
    \Mod{SFT} & \Mod{MCTS} & \Mod{36.96} & \Mod{53.84} & \Mod{25.24} & \Mod{42.80} & \Mod{42.92} & \Mod{29.72} & \Mod{47.52} & \Mod{39.86} \\
    \midrule
    DAPO             & Greedy sampling & 43.40 & 55.60 & 39.60 & 46.00 & 43.60 & 31.20 & 46.40 & 43.69 \\
    DAPO             & MCTS            & 45.48 & 56.64 & \textbf{42.48} & 45.96 & 40.04 & 31.44 & 48.00 & 44.29 \\
    DAPO + Entropy   & Greedy sampling & 44.60 & 57.60 & 41.00 & 44.60 & 44.60 & 31.40 & 48.00 & 44.54 \\
    DAPO + Entropy   & MCTS            & 46.08 & 56.56 & 40.68 & 46.08 & 40.08 & 32.00 & 48.96 & 44.35 \\
    DAPO + IMAGINE   & Greedy sampling & 44.40 & 56.60 & 40.60 & 44.20 & 42.40 & 32.20 & \underline{50.40} & 44.40 \\
    DAPO + IMAGINE   & MCTS            & \underline{46.80} & 56.52 & 40.12 & 46.04 & 39.56 & 32.44 & 47.52 & 44.14 \\
    \Mod{DAPO + MERCI} & \Mod{Greedy sampling} & \Mod{43.80} & \Mod{57.00} & \Mod{41.60} & \Mod{45.60} & \Mod{43.00} & \Mod{31.40} & \Mod{48.00} & \Mod{44.34} \\
    \Mod{DAPO + MERCI} & \Mod{MCTS} & \Mod{45.28} & \Mod{57.24} & \Mod{41.40} & \Mod{46.08} & \Mod{42.28} & \Mod{32.52} & \Mod{48.32} & \Mod{44.73} \\
    MCTS             & Greedy sampling & 42.40 & 58.00 & 41.60 & 46.20 & 43.00 & 31.60 & 46.40 & 44.17 \\
    MCTS             & MCTS            & 45.44 & 58.24 & 40.80 & 46.28 & 42.88 & 31.16 & 47.20 & 44.57 \\
    \midrule
    LPC-MCTS         & Greedy sampling & 44.40 & 57.00 & 42.00 & \underline{48.00} & 43.20 & \underline{33.40} & 48.80 & 45.26 \\
    LPC-MCTS         & MCTS            & 45.32 & \textbf{58.56} & \underline{42.32} & \textbf{48.20} & \underline{46.08} & 33.16 & 50.08 & \underline{46.25} \\
    \rowcolor{CornflowerBlue!12}
    LPC-MCTS & LPC-MCTS & \textbf{46.92} & \underline{58.48} & 42.12 & 47.96 & \textbf{46.88} & \textbf{34.24} & \textbf{51.20} & \textbf{46.83} \\
    \midrule[\heavyrulewidth]
    \multicolumn{10}{l}{\textit{Llama-3.2-3B-Instruct}} \\
    \addlinespace
    \Mod{SFT} & \Mod{Greedy sampling} & \Mod{39.40} & \Mod{51.60} & \Mod{26.20} & \Mod{39.40} & \Mod{35.20} & \Mod{27.60} & \Mod{47.20} & \Mod{38.09} \\
    \Mod{SFT} & \Mod{MCTS} & \Mod{38.12} & \Mod{52.40} & \Mod{26.88} & \Mod{40.16} & \Mod{35.68} & \Mod{25.76} & \Mod{45.92} & \Mod{37.85} \\
    \midrule
    DAPO             & Greedy sampling & 44.40 & 55.00 & 36.60 & 44.60 & 36.80 & 30.80 & 47.20 & 42.20 \\
    DAPO             & MCTS            & 44.52 & 55.92 & 37.92 & 45.08 & 39.04 & 30.32 & 46.24 & 42.72 \\
    DAPO + Entropy   & Greedy sampling & 45.60 & 55.00 & 40.00 & 45.00 & 38.40 & 31.40 & 46.40 & 43.11 \\
    DAPO + Entropy   & MCTS            & 46.52 & 55.80 & 40.80 & 45.68 & 37.68 & \textbf{32.00} & 45.92 & 43.49 \\
    DAPO + IMAGINE   & Greedy sampling & 44.00 & 55.20 & 40.80 & 44.80 & 40.00 & 30.40 & 47.20 & 43.20 \\
    DAPO + IMAGINE   & MCTS            & 46.32 & 55.76 & 40.72 & 44.56 & 39.28 & 31.40 & 47.52 & 43.65 \\
    \Mod{DAPO + MERCI} & \Mod{Greedy sampling} & \Mod{45.20} & \Mod{56.00} & \Mod{40.80} & \Mod{44.60} & \Mod{38.20} & \Mod{30.80} & \Mod{46.40} & \Mod{43.14} \\
    \Mod{DAPO + MERCI} & \Mod{MCTS} & \Mod{45.24} & \Mod{\underline{56.20}} & \Mod{41.48} & \Mod{44.68} & \Mod{40.92} & \Mod{30.76} & \Mod{46.72} & \Mod{43.71} \\
    MCTS             & Greedy sampling & 45.20 & 55.40 & 38.80 & 44.00 & \textbf{43.20} & 28.60 & 48.00 & 43.31 \\
    MCTS             & MCTS            & 45.28 & 56.04 & 39.80 & 44.56 & 40.28 & 30.04 & 48.32 & 43.47 \\
    \midrule
    LPC-MCTS         & Greedy sampling & 45.00 & \textbf{56.80} & \textbf{43.00} & 46.00 & 40.40 & 30.40 & \underline{48.80} & 44.34 \\
    LPC-MCTS         & MCTS            & \textbf{47.16} & 56.00 & \underline{41.88} & \underline{47.12} & 41.68 & 30.92 & 48.32 & \underline{44.73} \\
    \rowcolor{CornflowerBlue!12}
    LPC-MCTS & LPC-MCTS & \underline{46.80} & 56.16 & 41.84 & \textbf{47.40} & \underline{42.28} & \underline{31.72} & \textbf{50.08} & \textbf{45.18} \\
    \bottomrule
  \end{tabular}%
  }
  \end{center}
  \vspace{-0.6em}
\end{table*}

\paragraph{Mathematical reasoning.}
Table~\ref{tab:math_main} shows that LPC-MCTS achieves the highest average
performance on both backbones when used during both training and inference,
improving over standard MCTS by 4.96 points on Qwen2.5-7B and 4.37 points on
Qwen3-4B. Holding each LPC-MCTS-trained policy fixed, replacing discrete visit
counts with learned pseudo-counts at inference yields further gains of 3.79 and
2.71 points, isolating the benefit of LPC-MCTS during test-time search. Under
standard MCTS evaluation, LPC-MCTS-trained policies also outperform the
strongest alternative training method by 0.87 and 1.66 points. The consistent
improvements across both backbones and both stages provide strong empirical
support that mitigating count fragmentation benefits test-time exploration
and the collection of trajectories for policy optimization.

\paragraph{Search.}
Table~\ref{tab:search_main} shows improvements on search. Using
LPC-MCTS at both stages improves over standard MCTS by 2.26 points on
Qwen2.5-7B-Instruct and 1.71 points on Llama-3.2-3B-Instruct. Holding the
policy fixed, LPC-MCTS improves test-time MCTS by 0.58 and 0.45 points, while
LPC-MCTS training improves MCTS evaluation by 1.52 and 1.02 points over the
strongest alternatives. Compared with mathematical reasoning, however, these
gains are more modest. The logged trees provide one explanation: among 256
sampled trees per task, search covers 1.82 effective first-level branches on
average, compared with 4.05 for mathematical reasoning. This narrower structure
suggests that search is simpler in our setting and tends to follow more
standardized solution strategies, leaving fewer competing branches across which
pseudo-counts can share exploration evidence.

The ablations in Appendix~\ref{app:ablation} show that pseudo-count-guided
trajectory selection and token-to-node aggregation both improve performance,
while the sensitivity results in
Appendix~\ref{app:filtering_hyperparameters} favor $p=30$ on both backbones.
Overall, LPC-MCTS improves performance in both domains.
The larger gains in mathematical reasoning are consistent with richer branching, where
learned pseudo-counts can better distinguish novel paths and share exploration
evidence across related branches.
\Mod{An inference-time cost analysis is provided in Appendix~\ref{app:inference_cost}.}

\section{Conclusion}

We identify count fragmentation as a key limitation of MCTS for LLM reasoning: textually different branches can follow the same reasoning pattern while retaining independent visit counts.
We introduce LPC-MCTS, which replaces exact UCT visits with pseudo-counts from a lightweight CFN head and prioritizes less-covered trajectories for policy optimization.
Our theory quantifies the additional exploration cost caused by fragmented counts.
Across mathematical reasoning and search tasks, LPC-MCTS consistently improves over standard MCTS, while its learned counts align across related patterns and distinguish novel ones, showing that representation-level familiarity provides a more effective basis for exploration than textual branch identity.

\bibliography{iclr2027_conference}
\bibliographystyle{iclr2027_conference}

\appendix
\section{Addition descriptions for our method}
\label{app:implementation}

\subsection{Algorithmic Description of our method}
\label{app:method_algorithm}

The complete LPC-MCTS procedure is summarized in
Algorithm~\ref{alg:lpc_mcts}. The learned count enters the search policy, while
terminal outcomes provide the learning signal for policy optimization. The CFN
head is trained online and retained throughout policy training, sharing the
evidence accumulated across prompts.

\begin{algorithm}[t]
    \caption{LPC-MCTS training with learned pseudo-counts}
    \label{alg:lpc_mcts}
    \begin{algorithmic}[1]
        \REQUIRE Policy $\pi_\theta$, prompt dataset $\mathcal D$, verifier
        $R$, training iterations $N$, candidate limit $K_{\max}\geq B$,
        selected batch size $B$.
        \STATE Initialize CFN head $f_\phi$ randomly.
        \FOR{$i=1$ to $N$}
            \STATE Sample a prompt batch $\mathcal D_i$ from $\mathcal D$.
            \STATE Initialize the policy-training set
            $\mathcal D_{\mathrm{tree}}\leftarrow\varnothing$.
            \FOR{each prompt $x\in\mathcal D_i$}
                \STATE Initialize search-tree root $v_0$ and terminal pool
                $\mathcal C_x\leftarrow\varnothing$.
                \WHILE{$|\mathcal C_x|<K_{\max}$}
                    \STATE Select a path from $v_0$ using
                    Eq.~\ref{eq:learned_uct}, and expand its frontier node with
                    $\pi_{\theta_{\mathrm{old}}}$.
                    \STATE Continue tree expansion until obtaining a terminal
                    trajectory $\tau$.
                    \STATE During the rollout, construct decision nodes with
                    the entropy rule and compute their pseudo-counts using
                    Eq.~\ref{eq:node_pseudocount}.
                    \STATE Evaluate $R(\tau)$, update the tree statistics, and
                    add $\tau$ to $\mathcal C_x$.
                    \STATE Store the collected representations and newly
                    sampled Rademacher targets, and periodically update
                    $f_\phi$ using the newly collected samples.
                \ENDWHILE
                \STATE Select the $B$ least-covered trajectories from
                $\mathcal C_x$ using Eq.~\ref{eq:trajectory_selection}.
                \STATE Compute $b_x$ and $A(v,a)$ using
                Eq.~\ref{eq:tree_advantage}, and add the selected action
                segments to $\mathcal D_{\mathrm{tree}}$.
            \ENDFOR
            \STATE Update $\pi_\theta$ with the clipped objective in
            Eq.~\ref{eq:policy_objective}.
        \ENDFOR
        \ENSURE Optimized policy $\pi_\theta$ and online CFN head $f_\phi$.
    \end{algorithmic}
\end{algorithm}

\subsection{Policy Optimization Details}
\label{app:policy_optimization}

Unless stated otherwise, the following choices are shared by all compared
search variants. 

\paragraph{Bounded MCTS iterations.}
Rollout lengths and tool-interaction costs can vary substantially across
prompts, so unconstrained tree search may lead to excessively long rollouts
and unpredictable training time. To keep the computational cost bounded, we
set a maximum number of MCTS iterations, denoted by $T$, for each prompt. Once
this limit is reached, we stop expanding the search tree and proceed to policy
optimization using the trajectories collected so far. We set this practical
runtime budget to $T=24$ during training and $T=128$ during evaluation.

\paragraph{Task- and model-specific advantage adaptation.}
\label{app:advantage_adaptation}
For some model--task combinations, including Qwen2.5-7B on
tool-integrated mathematical reasoning, we observe that RL can weaken
tool-use behavior acquired during SFT. We therefore apply a small,
bounded format-aware adjustment to the policy advantage to encourage
valid tool use, with scale $\delta=0.1$. This training adaptation is
separate from the LPC-MCTS exploration rule.

\paragraph{Clipping and loss aggregation.}
Search trajectories are generated by $\pi_{\theta_{\mathrm{old}}}$ and then
reused to update $\pi_\theta$. As the two policies diverge, large importance
ratios can produce unstable gradients, especially for actions with negative
advantages. We control these updates with the asymmetric clipping ranges
$\epsilon_{\mathrm{low}}=0.2$ and $\epsilon_{\mathrm{high}}=0.28$ in
Eq.~\ref{eq:policy_objective}, together with a dual-clipping floor
$c_{\mathrm{dual}}=10$ for negative advantages \citep{ye2020mastering}.
Because trajectories have different lengths, we average the surrogate terms
over all trainable model-generated tokens in the batch so that longer
trajectories do not receive disproportionate weight. Tool-response tokens are
environment outputs rather than policy actions and are therefore excluded from
optimization. We use neither a KL penalty nor an entropy bonus.

\paragraph{Trajectory candidate selection.}
\label{app:terminal_pool}

LPC-MCTS assigns every completed trajectory in $\mathcal C_x$ the mean node
pseudo-count in Eq.~\ref{eq:trajectory_selection} and ranks the full candidate
pool in ascending order. All node scores used by this ranking are recorded
under the candidate pool's shared reference state. The first $B$ trajectories
form $\mathcal B_x$ and provide the trainable samples for policy optimization.
For tool-integrated mathematical reasoning, we set $K_{\max}=12$ and $B=8$;
for search task, we set $K_{\max}=8$ and $B=5$.

\subsection{Node Construction and Token Aggregation}
\label{app:node_construction}

As introduced in Section~\ref{sec:tree_policy}, $w$ denotes the number of
newly generated probe tokens used to decide whether to create a node after a
tool response. The coefficient $\alpha>1$ is the multiplicative threshold
that compares the probe window's mean token entropy with that of the preceding
context: a new node is created when the former exceeds $\alpha$ times the
latter. We use $w=16$ and $\alpha=2.0$.

\paragraph{Token filtering and aggregation.}
\label{app:token_aggregation}

The CFN encoder receives only the current node segment. Its score mask excludes
prompt, padding, and tool-response positions. Among eligible positions within
each node, we set $p=30$ and retain the top $p\%$ by $u_{v,t}$, then partition
their indices into maximal contiguous runs. Every run of at least two tokens
is kept, and $I_v$ is the union of the tokens in all such runs. If no run
reaches the minimum length, the highest-scoring eligible token is used as a
singleton fallback. We average the inverse-count scores of all tokens in
$I_v$ to obtain $U_\phi(v)$, then apply Eq.~\ref{eq:node_pseudocount}.

\subsection{CFN Training and Shared-Backbone Execution}
\label{app:cfn_execution}

Representations collected by the search worker are paired with Rademacher
targets and processed in vectorized minibatches. Eligibility masks are applied
before pairs enter the training set, so environment outputs are not encoded as
random-label training examples. The lightweight linear head is optimized with
the CFN regression objective and evaluated over token-level representations
collected within nodes in vectorized batches.

Each completed simulation contributes the eligible representations along its
selected path, paired with independently sampled Rademacher targets. New
samples are accumulated between updates and used to update the CFN head every
4 MCTS iterations. The head is initialized once at the beginning of policy
training; its parameters persist across prompts and policy-training steps. Its pseudo-counts therefore reflect the policy model's
accumulated evidence over the entire training process.

\paragraph{Shared-backbone execution.}
\label{app:shared_backbone}

The CFN and policy share the same LLM backbone and SGLang engine,
which avoids loading and
storing a separate backbone for CFN, and removes the need to swap
between separate policy and CFN backbone instances. During tree search, policy rollouts benefit from prefix KV-cache reuse.
CFN feature extraction performs a prefill over the node-local
segment to obtain the token-level hidden states required for scoring. Its additional cost merely consists of this feature extraction and lightweight head evaluation.

\section{Proof of the regret cost of fragmented counts}
\label{app:count_fragmentation_regret}

 We prove Theorem~\ref{theorem:count_fragmentation_regret} in this section.
Our finite-time bound adapts the classical UCB
analysis~\citep{auer2002finite}, while our asymptotic argument follows
the proof strategy of \citet{fan2022typical}.
We extend these arguments to shared exploration counts while retaining
separate action-level empirical means. This requires controlling
estimation errors using each action's own samples and establishing
consistency of the maximum empirical mean within each counting unit.

\subsection{Setup and a common form of the two algorithms}

We use the action groups defined in Section~\ref{sec:count_fragmentation_regret}:
each group $\mathcal C_c$ contains $m_c$ distinct actions with the same reward
distribution. Let $M=\sum_{c=1}^{K}m_c$ be the total number of actions.
For each $a\in\mathcal C_c$, let
$(X_{a,n})_{n\ge1}$ be its potential reward sequence, with
$X_{a,n}\sim P_c$, $P_c$ supported on $[0,1]$, and mean $\mu_c$.
All these reward variables are mutually independent. The $n$-th selection of
$a$ reveals $X_{a,n}$. Write
\[
    \overline X_{a,n}=\frac{1}{n}\sum_{i=1}^{n}X_{a,i},
    \qquad
    \widehat\mu_a(t)=\overline X_{a,N_a(t)}.
\]
Both algorithms select every action once during the first $M$ rounds and
then follow the UCB rules in Section~\ref{sec:count_fragmentation_regret}.
All counts and empirical means refer
to the respective algorithm's own history. The shared-count rule is given
the action grouping, but neither rule is given the reward means or the
identity of the optimal action. Each action's empirical mean uses only its
own observed rewards.

To analyze both rules together, call a set of actions that share one
exploration counter a \emph{counting unit}. Under independent-count UCB,
each action forms its own unit; under shared-count UCB, all actions in
$\mathcal C_c$ form one unit. Thus, an action group contains $m_c$ counting
units under the first rule and one under the second.
For a counting unit $g\subseteq\mathcal C_c$, define
\begin{equation}
    S_g(t)=\sum_{a\in g}N_a(t),
    \qquad
    G_g(t)=\max_{a\in g}\widehat\mu_a(t).
    \label{eq:cf_group_statistics}
\end{equation}
The largest index in $g$ is
\begin{equation}
    J_g(t)=G_g(t-1)+\sqrt{\frac{2\log t}{S_g(t-1)}}.
    \label{eq:cf_group_index}
\end{equation}
Thus, each algorithm selects a unit maximizing $J_g(t)$ and then an action
attaining $G_g(t-1)$. This is an exact rewriting of the action-level rule,
not a replacement of local values by a pooled estimate. Let
\[
    \rho_c^{\mathrm{ind}}=m_c,
    \qquad
    \rho_c^{\mathrm{share}}=1
\]
denote the number of counting units in action group $c$. The unique optimal
action, denoted $\star$, forms a unit by itself under both rules.

\subsection{A finite-time bound}

For $B\in\{\mathrm{ind},\mathrm{share}\}$, every $T\ge M$, every suboptimal
action group $c$, and every $\varepsilon\in(0,\Delta_c)$, we first show that
\begin{equation}
    \mathbb E_B[N_c(T)]
    \le m_c
       +\rho_c^B\left\lceil
          \frac{2\log T}{(\Delta_c-\varepsilon)^2}
        \right\rceil
       +\frac{m_c}{e^{2\varepsilon^2}-1}+2.
    \label{eq:cf_finite_count}
\end{equation}

At round $t>M$, let
\[
    E_t=\left\{
      \widehat\mu_\star(t-1)
      +\sqrt{\frac{2\log t}{N_\star(t-1)}}<\mu_1
    \right\}
\]
be the event that the optimal index underestimates its mean. By Hoeffding's
inequality~\citep{hoeffding1963probability} and a union bound over the
possible sample counts,
\begin{equation}
\begin{aligned}
    \Pr(E_t)
    &\le\sum_{n=1}^{t-1}
       \Pr\!\left(
         \overline X_{\star,n}
         <\mu_1-\sqrt{\frac{2\log t}{n}}
       \right)\\
    &\le(t-1)t^{-4}\le t^{-3}.
\end{aligned}
\label{eq:cf_optimal_underestimate}
\end{equation}
In particular, $\sum_{t=M+1}^{\infty}\Pr(E_t)<2$.
The union bound handles adaptive sampling without assuming that an empirical
mean is independent of its random sample count.

Fix $c$, $T$, and $\varepsilon$, and set
\[
    L_c=\left\lceil
       \frac{2\log T}{(\Delta_c-\varepsilon)^2}
    \right\rceil.
\]
Suppose action $a\in g\subseteq\mathcal C_c$ is selected at $t\le T$ with
$S_g(t-1)\ge L_c$. Its exploration bonus is at most
$\Delta_c-\varepsilon$. Outside $E_t$, its index must be at least $\mu_1$,
so
\begin{equation}
    \widehat\mu_a(t-1)\ge\mu_c+\varepsilon.
    \label{eq:cf_local_overestimate}
\end{equation}
Every such selection increases the selected action's own count. Therefore,
each pair $(a,n)$, where $n=N_a(t-1)$, can account for at most one selection.
The expected number of selections satisfying Eq.~\eqref{eq:cf_local_overestimate}
is consequently bounded by
\begin{equation}
\begin{aligned}
    \sum_{a\in\mathcal C_c}\sum_{n=1}^{T-1}
      \Pr(\overline X_{a,n}\ge\mu_c+\varepsilon)
    &\le m_c\sum_{n=1}^{\infty}e^{-2n\varepsilon^2}\\
    &=\frac{m_c}{e^{2\varepsilon^2}-1}.
\end{aligned}
\label{eq:cf_local_error_cost}
\end{equation}
There are $m_c$ initialization selections and at most $\rho_c^B L_c$
subsequent selections with the counting unit's count below $L_c$. Adding these terms and
the expected number of $E_t$ events proves Eq.~\eqref{eq:cf_finite_count}.
Crucially, every concentration bound uses the selected action's actual local
sample size; the shared count is never treated as the sample size of a local
empirical mean.

Taking $\varepsilon=\Delta_c/2$, using $e^x-1\ge x$ and
$\lceil x\rceil\le x+1$, and applying the regret decomposition in
Section~\ref{sec:count_fragmentation_regret} gives
\begin{equation}
    R_B(T)\le\sum_{c=2}^{K}\left[
      \frac{8\rho_c^B\log T}{\Delta_c}
      +\frac{2m_c}{\Delta_c}
      +(m_c+\rho_c^B+2)\Delta_c
    \right].
    \label{eq:cf_finite_regret}
\end{equation}
Hence, for each action group, the dependence improves from $m_c\log T$ to
$\log T+m_c$, up to gap-dependent factors. The additive dependence on $m_c$
accounts for initialization and errors in the separate empirical values.
Comparing these upper bounds alone would not establish an actual regret
separation; we prove the matching asymptotic statements next.

\subsection{Exact asymptotics}

\begin{proof}[Proof of Theorem~\ref{theorem:count_fragmentation_regret}]
\textbf{Every counting unit is selected infinitely often.}
Suppose some unit has a finite limiting count. Since its count is
integer-valued and nondecreasing, it eventually stops being selected. For
every unit, $G_g(t)\in[0,1]$, and
\[
    \frac{J_g(t)}{\sqrt{2\log t}}
    =\frac{G_g(t-1)}{\sqrt{2\log t}}
     +\frac{1}{\sqrt{S_g(t-1)}}.
\]
This expression has a positive limit for any unit with finite limiting
count, and tends to zero for a unit whose count diverges. There are finitely
many units, so all sufficiently late maximizing indices would belong to
units that have supposedly stopped being selected, a contradiction.
Thus $S_g(t)\to\infty$ for every unit.

\medskip
\noindent\textbf{The maximum local mean in each counting unit is consistent.}
Work on the probability-one event on which the strong law holds for every
action's potential reward sequence. In a unit $g\subseteq\mathcal C_c$, at
least one action is selected infinitely often, and the empirical mean of
every such action converges to $\mu_c$. Actions selected finitely often
eventually have frozen empirical means. No frozen mean can exceed $\mu_c$:
otherwise, after all finitely selected actions have received their last
samples and all infinitely selected actions have means below the largest
frozen mean, the next selection of $g$ would have to select a supposedly
frozen action. This contradicts the definition of those actions. Therefore,
\begin{equation}
    G_g(t)\longrightarrow\mu_c
    \qquad\text{almost surely}.
    \label{eq:cf_group_consistency}
\end{equation}
This argument permits some actions to receive only finitely many samples;
it relies on selecting a maximum-mean action whenever its unit is chosen.

\medskip
\noindent\textbf{Each suboptimal counting unit incurs one logarithmic sampling cost.}
Fix a unit $g\subseteq\mathcal C_c$, $c\ge2$, and
$\eta\in(0,\Delta_c/2)$. By Eq.~\eqref{eq:cf_group_consistency}, for all
sufficiently large $t$,
$G_g(t-1)\le\mu_c+\eta$ and $G_\star(t-1)\ge\mu_1-\eta$.
If $g$ is selected at such a round, comparison with the optimal index gives
\[
    \mu_c+\eta+\sqrt{\frac{2\log t}{S_g(t-1)}}
    \ge\mu_1-\eta,
    \qquad
    S_g(t-1)\le\frac{2\log t}{(\Delta_c-2\eta)^2}.
\]
Applying this inequality at the last selection of $g$ up to $T$ yields
\begin{equation}
    \limsup_{T\to\infty}\frac{S_g(T)}{\log T}
    \le\frac{2}{(\Delta_c-2\eta)^2}
    \qquad\text{almost surely}.
    \label{eq:cf_pathwise_upper}
\end{equation}
Since the number of suboptimal counting units is finite,
$N_\star(T)=T-O(\log T)$ almost surely.

Let $\tau(T)$ be the last round up to $T$ at which the optimal action is
selected. Every subsequent round is suboptimal, so
$\tau(T)=T-O(\log T)$ and $\log\tau(T)/\log T\to1$.
At round $\tau=\tau(T)$, the optimal index dominates $J_g(\tau)$, and hence
\begin{equation}
    \sqrt{\frac{2\log\tau}{S_g(\tau-1)}}
    \le G_\star(\tau-1)-G_g(\tau-1)
      +\sqrt{\frac{2\log\tau}{N_\star(\tau-1)}}
    =\Delta_c+o(1).
    \label{eq:cf_pathwise_lower_comparison}
\end{equation}
The final bonus tends to zero because the optimal action receives
$\tau-O(\log\tau)$ samples. Using $S_g(T)\ge S_g(\tau-1)$ therefore gives
\[
    \liminf_{T\to\infty}\frac{S_g(T)}{\log T}
    \ge\frac{2}{\Delta_c^2}.
\]
Letting $\eta\downarrow0$ through a countable sequence in
Eq.~\eqref{eq:cf_pathwise_upper} proves
\begin{equation}
    \frac{S_g(T)}{\log T}\longrightarrow\frac{2}{\Delta_c^2},
    \qquad
    \frac{N_c^B(T)}{\log T}\longrightarrow
      \frac{2\rho_c^B}{\Delta_c^2}
    \qquad\text{almost surely}.
    \label{eq:cf_pathwise_limits}
\end{equation}

\medskip
\noindent\textbf{Passing to expected regret.}
For each fixed $\varepsilon\in(0,\Delta_c)$,
Eq.~\eqref{eq:cf_finite_count} implies
\[
    \limsup_{T\to\infty}
      \frac{\mathbb E_B[N_c(T)]}{\log T}
    \le\frac{2\rho_c^B}{(\Delta_c-\varepsilon)^2}.
\]
Letting $\varepsilon\downarrow0$ bounds this limsup by
$2\rho_c^B/\Delta_c^2$. Conversely, Fatou's lemma and
Eq.~\eqref{eq:cf_pathwise_limits} yield
\[
    \liminf_{T\to\infty}
      \frac{\mathbb E_B[N_c(T)]}{\log T}
    \ge\mathbb E_B\!\left[
      \liminf_{T\to\infty}\frac{N_c(T)}{\log T}
    \right]
    =\frac{2\rho_c^B}{\Delta_c^2}.
\]
Thus the expected counts have the same limiting coefficients. Substitution
into the regret decomposition in Section~\ref{sec:count_fragmentation_regret},
with $\rho_c^{\mathrm{ind}}=m_c$ and $\rho_c^{\mathrm{share}}=1$, proves the theorem.
\end{proof}

\paragraph{The resulting separation.}
The theorem compares actual expected regrets, giving
\begin{equation}
    R_{\mathrm{ind}}(T)-R_{\mathrm{share}}(T)
    =2\log T\sum_{c=2}^{K}\frac{m_c-1}{\Delta_c}+o(\log T),
    \label{eq:cf_regret_difference}
\end{equation}
and
\begin{equation}
    \lim_{T\to\infty}
      \frac{R_{\mathrm{ind}}(T)}{R_{\mathrm{share}}(T)}
    =\frac{\sum_{c=2}^{K}m_c/\Delta_c}
           {\sum_{c=2}^{K}1/\Delta_c}.
    \label{eq:cf_regret_ratio}
\end{equation}
If every suboptimal action group contains $m$ actions, this ratio is $m$. Deterministic
reward distributions satisfy the assumptions, so the separation can occur
even when all local empirical values are exact after initialization.

\section{Count Fragmentation in Recursive Tree Search}
\label{app:tree_count_fragmentation}

We extend the bandit analysis to recursive UCT with terminal-reward backup.
Our proof follows the recursive strategy used in UCT
analysis~\citep{kocsis2006bandit} and the subtree-regret decomposition
of \citet{coquelin2007bandit}: we control the error in parent-edge value
estimates through the regret accumulated in child subtrees.
We adapt this strategy to shared exploration counts while retaining
separate empirical action values. In the fixed deterministic tree
considered here, this yields matching regret asymptotics that quantify
the cost of count fragmentation along the optimal path.

\paragraph{Tree setting.}
Consider a fixed finite rooted tree $\mathcal T$ with root $v_0$.
Each leaf $\ell$ has a deterministic reward $r_\ell\in[0,1]$, and
$V^*(v)$ denotes the largest leaf reward reachable from node $v$.
At each internal node $v$, divide the actions into groups
$\mathcal C_{v,c}$, with $m_{v,c}=|\mathcal C_{v,c}|$ actions in each group.
Actions in the same group lead to separate subtrees with identical
structure and matching leaf rewards.
Writing $v_a$ for the child reached by action $a$,
their common optimal continuation value is
$\mu_{v,c}=V^*(v_a)$ for $a\in\mathcal C_{v,c}$.

Assume every internal node has a unique optimal action $a^*(v)$,
whose group, indexed by $c^*(v)$, contains only that action.
For each suboptimal action group, define
$\Delta_{v,c}=V^*(v)-\mu_{v,c}>0$.
Let $P^*$ denote the internal nodes on the unique optimal
root-to-leaf path.

\paragraph{Selection and backup.}
Each simulation recursively selects a root-to-leaf path and backs up
the terminal reward along that path. Let $N(v)$ and $N(v,a)$ be the
numbers of completed simulations through node $v$ and edge $(v,a)$,
respectively, and define
$N(v,c)=\sum_{a\in\mathcal C_{v,c}}N(v,a)$.
The empirical value $\widehat Q(v,a)$ averages only the terminal
rewards observed through edge $(v,a)$.

At each node, both algorithms first try every action once and then
select an action maximizing the corresponding index:
\begin{equation}
\begin{aligned}
    I_{\mathrm{ind}}(v,a)
    &=\widehat Q(v,a)
      +\sqrt{\frac{2\log N(v)}{N(v,a)}},\\
    I_{\mathrm{share}}(v,a)
    &=\widehat Q(v,a)
      +\sqrt{\frac{2\log N(v)}{N(v,c(a))}}.
\end{aligned}
\label{eq:tree_cf_indices}
\end{equation}
Thus, both algorithms retain separate empirical action values and
use the same terminal-reward backup procedure; only their exploration
counts differ.

Following the cumulative-regret formulation for
tree search~\citep{coquelin2007bandit}, define
\begin{equation}
    \mathcal R(T)
    =\sum_{t=1}^{T}\bigl(V^*(v_0)-r_{\ell_t}\bigr),
    \label{eq:tree_cf_regret}
\end{equation}
where $\ell_t$ is the leaf reached by simulation $t$.

\begin{theorem}[Regret penalty of count fragmentation in trees]
\label{prop:tree_count_fragmentation}
Under the setting above, for the fixed tree as $T\to\infty$,
\begin{equation}
\begin{aligned}
    \mathcal R_{\mathrm{ind}}(T)
    &=2\log T
      \sum_{v\in P^*}\sum_{c\ne c^*(v)}
      \frac{m_{v,c}}{\Delta_{v,c}}
      +O_{\mathcal T}(\log\log T),\\
    \mathcal R_{\mathrm{share}}(T)
    &=2\log T
      \sum_{v\in P^*}\sum_{c\ne c^*(v)}
      \frac{1}{\Delta_{v,c}}
      +O_{\mathcal T}(\log\log T).
\end{aligned}
\label{eq:tree_cf_separation}
\end{equation}
If there is at least one suboptimal action group on $P^*$ and every such
group contains $m$ actions, then
$\mathcal R_{\mathrm{ind}}(T)/
 \mathcal R_{\mathrm{share}}(T)\to m$.
The constants in the remainder may depend on the fixed tree.
\end{theorem}

\begin{proof}
We use local visit counts and proceed by induction on subtree height.
Each subtree is analyzed in its own local visit time. Its search process
can be continued beyond the visits made by the parent, so the bounds
also apply to any finite observed prefix.

\paragraph{Recursive identities.}
For a node $v$, let $n$ denote its number of visits and let
$N_{v,a}(n)$ count selections of action $a$ during those visits.
Define the local subtree regret as
\begin{equation}
    \mathcal R_v(n)
    =nV^*(v)-\sum_{i=1}^{n}r_{\ell_i^{(v)}},
    \qquad
    \mathcal R_v(0)=0,
    \label{eq:tree_cf_local_regret}
\end{equation}
where $\ell_i^{(v)}$ is the leaf reached on the $i$th visit to $v$.
For a leaf, $\mathcal R_v(n)=0$.

Since every visit to child $v_a$ passes through edge $(v,a)$,
the empirical edge value satisfies, for $k=N_{v,a}(n)>0$,
\begin{equation}
    \widehat Q_v(a;n)
    =V^*(v_a)-\frac{\mathcal R_{v_a}(k)}{k}.
    \label{eq:tree_cf_value_identity}
\end{equation}
Thus, the error in the parent edge value is exactly the average
regret accumulated within the child subtree.

Writing $\Delta(v,a)=V^*(v)-V^*(v_a)$, we also have
\begin{equation}
    \mathcal R_v(n)
    =\sum_{a\in\mathcal A(v)}\Delta(v,a)N_{v,a}(n)
     +\sum_{a\in\mathcal A(v)}
       \mathcal R_{v_a}\!\left(N_{v,a}(n)\right).
    \label{eq:tree_cf_recursion}
\end{equation}
This follows by decomposing each simulation's regret as
$V^*(v)-r_\ell
 =[V^*(v)-V^*(v_a)]+[V^*(v_a)-r_\ell]$.

\paragraph{Induction hypothesis and counting units.}
Suppose every child subtree $u$ satisfies
\begin{equation}
    \mathcal R_u(k)
    =A_u\log k+O_{\mathcal T}(\log\log k)
    \qquad\text{as }k\to\infty.
    \label{eq:tree_cf_induction}
\end{equation}
In particular, for a suitable finite constant $C_u$,
\begin{equation}
    0\le\mathcal R_u(k)\le C_u\log(k+\mathrm e)
    \qquad\text{for all }k\ge0.
    \label{eq:tree_cf_child_bound}
\end{equation}
These statements hold at leaves with $A_u=C_u=0$.

At the parent node $v$, a \emph{counting unit} is a set of actions
sharing one exploration counter. Under independent-count UCT, each
action forms its own unit; under shared-count UCT, all actions in
$\mathcal C_{v,c}$ form one unit. For a counting unit $g$, suppressing
$v$ in the notation, define
\begin{equation}
\begin{aligned}
    S_g(n)&=\sum_{a\in g}N_{v,a}(n),\\
    G_g(n)&=\max_{a\in g}\widehat Q_v(a;n),\\
    J_g(n)&=G_g(n)+\sqrt{\frac{2\log n}{S_g(n)}}.
\end{aligned}
\label{eq:tree_cf_groups}
\end{equation}
Maximizing the action indices is equivalent to choosing a unit
maximizing $J_g(n)$ and then an action attaining $G_g(n)$.
This is only a rewriting of the selection rule: empirical values
remain separate.

Let $\mu_g$ be the common optimal child value within unit $g$,
and let $\Delta_g=V^*(v)-\mu_g$.
The optimal counting unit, denoted by $g_*$, contains only the optimal action.

\paragraph{Consistency and a coarse count bound.}
Every counting unit is visited infinitely often. Otherwise, some
units would have eventually constant positive counts $s_g$.
For such units,
$J_g(n)/\sqrt{2\log n}\to1/\sqrt{s_g}>0$,
whereas this normalized index tends to zero for units whose
counts diverge. Since there are finitely many units, a maximizing
unit would eventually have to belong to the set whose counts
supposedly stopped increasing, a contradiction.

We next control the value estimate within each counting unit.
By Eq.~\eqref{eq:tree_cf_value_identity},
$\widehat Q_v(a;n)\le\mu_g$ for every $a\in g$.
At least one action in $g$ has count $k\ge S_g(n)/|g|$.
Applying Eq.~\eqref{eq:tree_cf_child_bound} to that action gives
\begin{equation}
\begin{aligned}
    0\le b_g(n)
    &:=\mu_g-G_g(n)\\
    &\le D_g\,
      \frac{\log(S_g(n)+\mathrm e)}{S_g(n)},
\end{aligned}
\label{eq:tree_cf_group_bias}
\end{equation}
where $D_g=|g|\max_{a\in g}C_{v_a}$.
Hence $G_g(n)\to\mu_g$.
This argument does not require every action within a shared counting
unit to be visited infinitely often.

If a suboptimal unit $g$ is selected after $n$ completed visits,
then
\[
    \mu_g+\sqrt{\frac{2\log n}{S_g(n)}}
    \ge J_g(n)\ge J_*(n)\ge G_*(n).
\]
For sufficiently large $n$, consistency gives
$G_*(n)\ge V^*(v)-\Delta_g/2$, and therefore
$S_g(n)\le8\log n/\Delta_g^2$ at such selection times.
Applying this bound at the last selection of $g$ and allowing
for its unit increment yields $S_g(n)=O_{\mathcal T}(\log n)$
at every sufficiently large time. Consequently,
\begin{equation}
    S_*(n)=n-O_{\mathcal T}(\log n),
    \qquad
    b_*(n)=O_{\mathcal T}\!\left(\frac{\log n}{n}\right).
    \label{eq:tree_cf_optimal_count_bias}
\end{equation}

\paragraph{Matching upper and lower count bounds.}
We first sharpen the upper bound.
Whenever a suboptimal unit $g$ is selected,
comparison with the optimal unit gives
\[
    \sqrt{\frac{2\log n}{S_g(n)}}
    \ge\Delta_g-b_*(n).
\]
Using Eq.~\eqref{eq:tree_cf_optimal_count_bias} and the fixed
positive gap $\Delta_g$,
\[
    S_g(n)
    \le\frac{2\log n}{(\Delta_g-b_*(n))^2}
    =\frac{2\log n}{\Delta_g^2}
      +O_{\mathcal T}\!\left(\frac{(\log n)^2}{n}\right).
\]
Applying this inequality at the last selection of $g$, including
its unit increment, proves
\begin{equation}
    S_g(n)\le\frac{2\log n}{\Delta_g^2}
                    +O_{\mathcal T}(1).
    \label{eq:tree_cf_count_upper}
\end{equation}

For the lower bound, let $\tau\le n$ be the last visit on which
the optimal action was selected, and put $s=\tau-1$.
All visits after $\tau$ are suboptimal, so the coarse count bound
implies
$s=n-O_{\mathcal T}(\log n)$ and $\log s=\log n+o(1)$.
At the decision after $s$ completed visits, $J_*(s)\ge J_g(s)$.
Since $G_*(s)\le V^*(v)$,
\begin{equation}
    \sqrt{\frac{2\log s}{S_g(s)}}
    \le\Delta_g+b_g(s)
       +\sqrt{\frac{2\log s}{S_*(s)}}.
    \label{eq:tree_cf_lower_comparison}
\end{equation}
The last two terms tend to zero, so
$S_g(s)\ge c_g\log s$ for some constant $c_g>0$.
Together with Eq.~\eqref{eq:tree_cf_count_upper}, this gives
$S_g(s)=\Theta_{\mathcal T}(\log s)$ at these comparison times.
Equation~\eqref{eq:tree_cf_group_bias} then yields
\[
    b_g(s)
    =O_{\mathcal T}\!\left(\frac{\log\log s}{\log s}\right),
    \qquad
    \sqrt{\frac{2\log s}{S_*(s)}}
    =O_{\mathcal T}\!\left(\sqrt{\frac{\log s}{s}}\right).
\]
Substituting these estimates into
Eq.~\eqref{eq:tree_cf_lower_comparison} and expanding the
reciprocal square around the fixed gap $\Delta_g$ gives
\[
    S_g(s)\ge
    \frac{2\log s}{\Delta_g^2}
    -O_{\mathcal T}(\log\log s).
\]
Indeed, the first error contributes $O(\log\log s)$ after
multiplication by $\log s$, while the second contributes
$O((\log s)^{3/2}/\sqrt{s})=o(1)$.
Using $S_g(n)\ge S_g(s)$ and $\log s=\log n+o(1)$ proves
\begin{equation}
    S_g(n)\ge\frac{2\log n}{\Delta_g^2}
                    -O_{\mathcal T}(\log\log n).
    \label{eq:tree_cf_count_lower}
\end{equation}
Combining the two bounds,
\begin{equation}
    S_g(n)
    =\frac{2\log n}{\Delta_g^2}
       +O_{\mathcal T}(\log\log n).
    \label{eq:tree_cf_threshold}
\end{equation}

\paragraph{From local counts to tree regret.}
For algorithm $B\in\{\mathrm{ind},\mathrm{share}\}$, define
$\rho^{\mathrm{ind}}_{v,c}=m_{v,c}$ and
$\rho^{\mathrm{share}}_{v,c}=1$.
Independent counting has one counting unit per action, whereas
shared counting has one per action group. Summing
Eq.~\eqref{eq:tree_cf_threshold} within each suboptimal action group gives
\begin{equation}
    N^B_{v,c}(n)
    =\frac{2\rho^B_{v,c}\log n}{\Delta_{v,c}^2}
       +O_{\mathcal T}(\log\log n).
    \label{eq:tree_cf_class_count}
\end{equation}
The local action-gap contribution in
Eq.~\eqref{eq:tree_cf_recursion} is therefore
\[
    2\log n\sum_{c\ne c^*(v)}
        \frac{\rho^B_{v,c}}{\Delta_{v,c}}
    +O_{\mathcal T}(\log\log n).
\]

Let $u_*=v_{a^*(v)}$ be the optimal child.
It receives $n_*=n-O_{\mathcal T}(\log n)$ visits, so the
induction hypothesis gives
\[
    \mathcal R^B_{u_*}(n_*)
    =A^B_{u_*}\log n+O_{\mathcal T}(\log\log n).
\]
Every suboptimal child receives at most
$O_{\mathcal T}(\log n)$ visits. Its additional subtree regret
is bounded by
\[
    \mathcal R^B_{v_a}\!\left(N^B_{v,a}(n)\right)
    \le C_{v_a}\log\!\left(N^B_{v,a}(n)+\mathrm e\right)
    =O_{\mathcal T}(\log\log n).
\]
This bound also covers actions that receive only finitely many visits.

Substitution into Eq.~\eqref{eq:tree_cf_recursion} proves
\[
    \mathcal R^B_v(n)
    =A^B_v\log n+O_{\mathcal T}(\log\log n),
\]
where
\begin{equation}
    A^B_v
    =A^B_{u_*}
     +2\sum_{c\ne c^*(v)}
       \frac{\rho^B_{v,c}}{\Delta_{v,c}}.
    \label{eq:tree_cf_coefficient}
\end{equation}
This completes the induction.

Finally, unrolling Eq.~\eqref{eq:tree_cf_coefficient} along the
unique optimal path yields
\[
    A^B_{v_0}
    =2\sum_{v\in P^*}\sum_{c\ne c^*(v)}
       \frac{\rho^B_{v,c}}{\Delta_{v,c}}.
\]
Taking $\rho^{\mathrm{ind}}_{v,c}=m_{v,c}$ and
$\rho^{\mathrm{share}}_{v,c}=1$ proves
Eq.~\eqref{eq:tree_cf_separation}.
When every suboptimal action group on $P^*$ contains $m$ actions, the independent-count
coefficient is $m$ times the shared-count coefficient, giving the
stated regret ratio.
\end{proof}

\section{Prompt Template and Reward Construction}
\label{app:prompt_and_reward}

For both settings, the task instance is placed in the user message, while the
prompt specifies the available tool and required function-call format. For
each backbone, the corresponding prompt is identical for DAPO, standard MCTS,
and LPC-MCTS.

\paragraph{Tool-integrated mathematical reasoning.}
We use the following system prompt with the \texttt{code\_interpreter} tool:

\begin{promptbox}{Mathematical-reasoning system prompt}
You are Qwen, created by Alibaba Cloud. You are a helpful assistant.

# Tools

You may call one or more functions to assist with the user query.

You are provided with function signatures within <tools></tools> XML tags:
<tools>
{"type": "function",
 "function": {
   "name": "code_interpreter",
   "description": "A tool for executing code.",
   "parameters": {
     "type": "object",
     "properties": {
       "code": {"type": "string", "description": "The code to execute."}
     },
     "required": ["code"]
   }
 }}
</tools>

For each function call, return a JSON object with function name and arguments within <tool_call></tool_call> XML tags:
<tool_call>
{"name": <function-name>, "arguments": <args-json-object>}
</tool_call>
\end{promptbox}

\paragraph{Search.}
For Qwen2.5-7B-Instruct, we use the following system prompt with the
\texttt{search} tool:

\begin{promptbox}{Qwen search system prompt}
You are Qwen, created by Alibaba Cloud. You are a helpful assistant.

# Tools

You may call one or more functions to assist with the user query.

You are provided with function signatures within <tools></tools> XML tags:
<tools>
{"type": "function",
 "function": {
   "name": "search",
   "description": "Searches the web for relevant information based on the given query.",
   "parameters": {
     "type": "object",
     "properties": {
       "query_list": {
         "type": "array",
         "items": {"type": "string"},
         "description": "A list of fully-formed semantic queries. The tool will return search results for each query."
       }
     },
     "required": ["query_list"]
   }
 }}
</tools>

For each function call, return a JSON object with function name and arguments within <tool_call></tool_call> XML tags:
<tool_call>
{"name": <function-name>, "arguments": <args-json-object>}
</tool_call>
\end{promptbox}

For Llama-3.2-3B-Instruct, the serving framework automatically adds the system
metadata at runtime. We therefore show only the user-message template below:

\begin{promptbox}{Llama search user prompt}
Given the following functions, please respond with a JSON for a function call with its proper arguments that best answers the given prompt.

Respond in the format {"name": function name, "parameters": dictionary of argument name and its value}. Do not use variables.

{
    "type": "function",
    "function": {
        "name": "search",
        "description": "Searches the web for relevant information based on the given query.",
        "parameters": {
            "type": "object",
            "properties": {
                "query_list": {
                    "type": "array",
                    "description": "A list of fully-formed semantic queries. The tool will return search results for each query.",
                    "enum": null
                }
            },
            "required": [
                "query_list"
            ]
        },
        "strict": false
    }
}
\end{promptbox}

The model can alternate between reasoning and tool calls, using each tool
output to continue its response. Because tool outputs are generated by the
environment rather than the policy, their tokens are excluded from both the
policy and CFN losses.

During training, we extract the final answer from
\texttt{\textbackslash boxed\{\}} for both tasks. For mathematical reasoning,
we normalize the extracted expression before comparing it with the reference
answer; for search, we use exact match against the reference answer. We define
the base task score as
\begin{equation}
S_{\text{base}} =
\begin{cases}
+1, & \text{if the final answer is correct}, \\
-1, & \text{otherwise}.
\end{cases}
\end{equation}

To address potential tool-call collapse in some RL settings, we add an
auxiliary tool-call reward to DAPO training. For incorrect trajectories, the bonus
increases by $0.05$ for each interaction turn beyond the first two and is
capped at $0.4$. Correct
trajectories receive a unit bonus when they contain more than two turns.

\section{Training and Inference Configurations}
\label{app:training_configuration}

\subsection{Training Configurations}

All experiments were conducted on 8 GPUs. Tool-integrated
mathematical reasoning uses Qwen2.5-7B and Qwen3-4B, while search uses
Qwen2.5-7B-Instruct and Llama-3.2-3B-Instruct.

\paragraph{Supervised Fine-Tuning.}
Before RL training, we perform SFT on each policy backbone. For
tool-integrated mathematical reasoning, Qwen2.5-7B and Qwen3-4B are trained on
the ReTool SFT corpus \citep{feng2026retool}. For search, Qwen2.5-7B-Instruct
and Llama-3.2-3B-Instruct are trained on the same 2,000 examples sampled from
the Chain-of-Agents SFT corpus
\citep{li2025chainofagentsendtoendagentfoundation}. The SFT configurations are
summarized in Table~\ref{tab:sft_configuration}.

\begin{table}[t]
  \caption{Supervised fine-tuning configurations.}
  \label{tab:sft_configuration}
  \begin{center}
  \begin{tabular}{ll}
    \toprule
    Hyperparameter & Value \\
    \midrule
    Optimizer & AdamW \\
    Learning rate & $1\times10^{-5}$ \\
    Weight decay & 0.01 \\
    Learning-rate schedule & Cosine \\
    Warmup ratio & 0.1 \\
    Global batch size & 16 \\
    Maximum sequence length & 16,384 (math) / 8,192 (search) \\
    Training epochs & 2 \\
    \bottomrule
  \end{tabular}
  \end{center}
\end{table}

\paragraph{Policy training.}

Tables~\ref{tab:math_policy_training_configuration}
and~\ref{tab:search_policy_training_configuration} summarize the policy
training settings for the two tasks. We train each experiment for 100 steps.

\begin{table}[t]
  \caption{Policy training configuration for tool-integrated mathematical
  reasoning.}
  \label{tab:math_policy_training_configuration}
  \begin{center}
  \begin{tabular}{ll}
    \toprule
    Hyperparameter & Value \\
    \midrule
    \multicolumn{2}{l}{\textit{General training hyperparameters}} \\
    Backbones & Qwen2.5-7B / Qwen3-4B \\
    Optimizer & AdamW \\
    Policy learning rate & $1\times10^{-6}$ \\
    Prompts per training batch & 128 \\
    Samples per prompt & 8 \\
    Mini-batch size & 128 \\
    Max prompt length & 1,024 \\
    Max response length & 18,000 \\
    Rollout temperature & 1.0 \\
    \midrule
    \multicolumn{2}{l}{\textit{MCTS hyperparameters}} \\
    Maximum tree depth & 6 \\
    Maximum tool-interaction turns per node & 3 \\
    UCT coefficient $c_{\mathrm{count}}$ & 2.0 \\
    Maximum response length per node & 3,072 \\
    \bottomrule
  \end{tabular}
  \end{center}
\end{table}

\begin{table}[t]
  \caption{Policy training configuration for search.}
  \label{tab:search_policy_training_configuration}
  \begin{center}
  \begin{tabular}{ll}
    \toprule
    Hyperparameter & Value \\
    \midrule
    \multicolumn{2}{l}{\textit{General training hyperparameters}} \\
    Backbones & Qwen2.5-7B-Instruct / Llama-3.2-3B-Instruct \\
    Optimizer & AdamW \\
    Policy learning rate & $1\times10^{-6}$ \\
    Prompts per training batch & 512 \\
    Samples per prompt & 5 \\
    Mini-batch size & 256 \\
    Max prompt length & 512 \\
    Max response length & 8,192 \\
    Rollout temperature & 1.0 \\
    \midrule
    \multicolumn{2}{l}{\textit{MCTS hyperparameters}} \\
    Maximum tree depth & 4 \\
    Maximum tool-interaction turns per node & 2 \\
    UCT coefficient $c_{\mathrm{count}}$ & 2.0 \\
    Maximum response length per node & 2048 \\
    \bottomrule
  \end{tabular}
  \end{center}
\end{table}

\paragraph{CFN configuration.}
\label{app:cfn_configuration_section}

Table~\ref{tab:cfn_configuration} summarizes the parameters specific to the
CFN estimator. The CFN head is randomly initialized and trained online. During
search, newly collected samples are accumulated, and the head is updated after
every four MCTS iterations. Backbone representations are treated as fixed
inputs to the CFN loss, so these updates do not propagate gradients into the
policy backbone.

\begin{table}[t]
  \caption{CFN estimator and online-training configurations.}
  \label{tab:cfn_configuration}
  \begin{center}
  \begin{tabular}{ll}
    \toprule
    Hyperparameter & Value \\
    \midrule
    CFN output dimension $d$ & 20 \\
    Optimizer & AdamW \\
    Online learning rate & $1\times10^{-4}$ \\
    CFN-head update frequency & Every 4 MCTS iterations \\
    Token filtering & Top 30\% \\
    \bottomrule
  \end{tabular}
  \end{center}
\end{table}

\paragraph{Training cost.}
On 8 GPUs, a full training run takes approximately 100 hours for tool-integrated mathematical reasoning and 50 hours for
search. Across all evaluated tasks and backbones, LPC-MCTS increases training
time by no more than 20\% relative to standard MCTS, making it a lightweight
addition for exploration through learning the pseudo-count.

\subsection{Inference Configurations}
\label{app:inference_configuration}
For sampling-based evaluation of mathematical reasoning, we use a temperature
of $0.6$ and top-$p$ sampling with $p=0.95$. For search, we evaluate greedy decoding and tree-based
inference with standard MCTS or LPC-MCTS with a temperature of $0.6$ and
top-$p$ sampling with $p=0.95$. \Mod{For MCTS-based evaluation, we run
a search tree for each problem and collect 16 terminal trajectories for mathematical reasoning or 5 for search. Tree search stops once the required number of trajectories is collected or the maximum of 128 MCTS iterations is reached.  If the limit is reached before collecting enough trajectories, each missing candidate is counted as incorrect. We report mean@16 (mathematical reasoning) or mean@5 (search) as the fraction of these trajectories whose final
answers are correct}, using the same protocol and search budget for standard
MCTS and LPC-MCTS.

\subsection{Search Retrieval Configuration}
\label{app:retrieval_configuration}

All search experiments use the same offline retrieval service. We use
\texttt{PeterJinGo/wiki-18-corpus}, a 2018 Wikipedia corpus containing
21,015,324 documents. Queries are encoded on CPU with
\texttt{intfloat/e5-base-v2} into 768-dimensional embeddings.

Retrieval uses the precomputed
\texttt{PeterJinGo/wiki-18-e5-index-HNSW64} FAISS HNSW64 index, which contains
21,015,324 vectors and uses inner-product similarity. We set
\texttt{efSearch} to 128 and return the top three documents for each query.

\section{Additional Experimental Results}
\label{app:add_results}
\subsection{Coin Flipping Network}
\label{app:add_results_cfn}

\subsubsection{Token-level Uncertainty Visualization}
\label{app:cfn_token_visualization}
For each selected response, we align the raw scalar score $u_{v,t}$ with every
eligible token position and preserve the original generation order.
Prompt, padding, and tool-response positions are excluded using the same mask
as in the node-level estimator. A larger $u_{v,t}$ indicates higher CFN uncertainty due to less accumulated
representation evidence.
\begin{figure}[!htbp]
  \centering
  \includegraphics[width=0.87\textwidth]{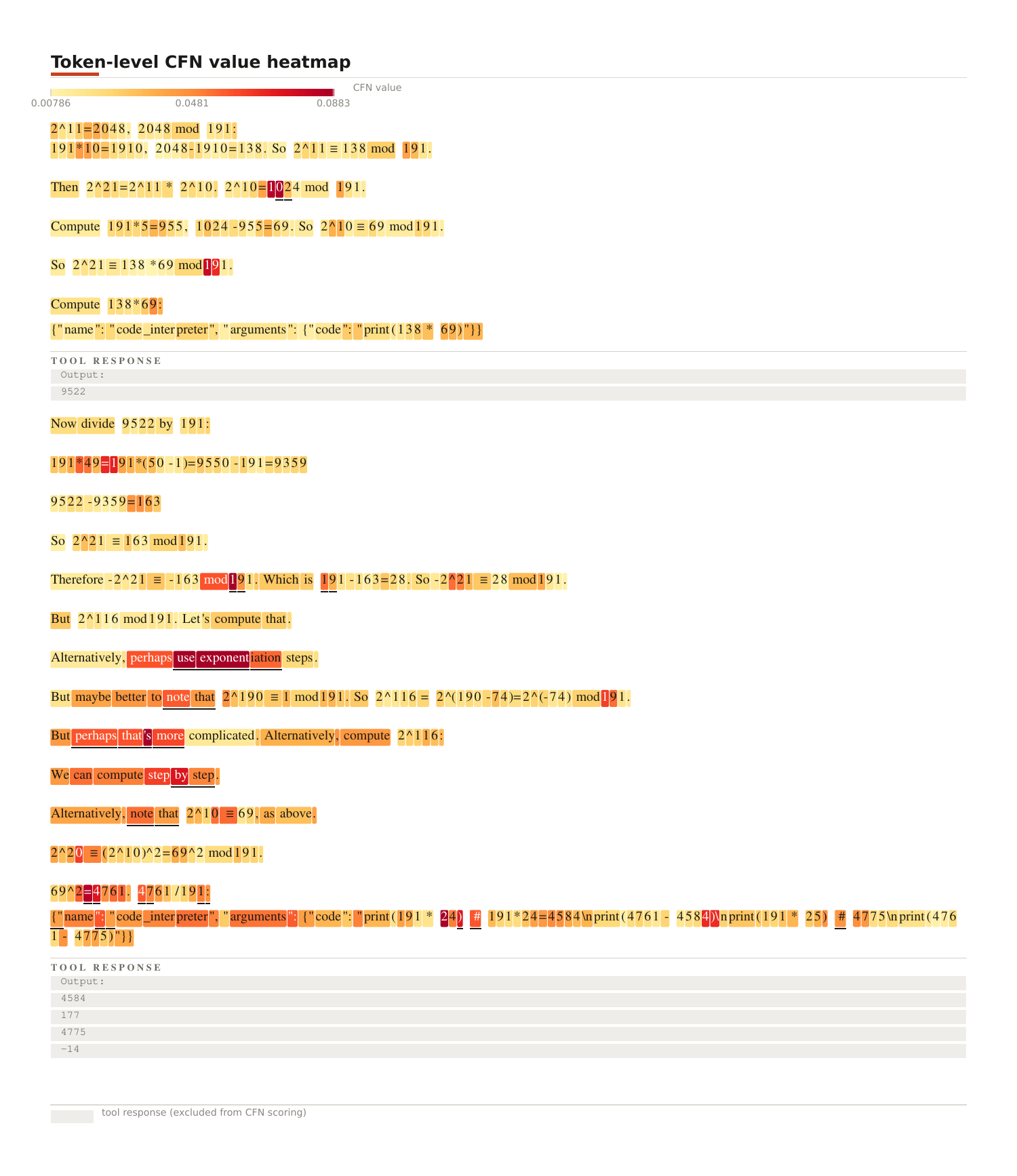}
  \caption{\textbf{An example of token-level uncertainty estimated by CFN within a Qwen3-4B modular-arithmetic response.}
  Each eligible generated token is colored by its raw CFN score $u_{v,t}$. Darker colors indicate higher uncertainty estimate, while lighter colors indicate relatively lower estimates. 
  Tool responses are shown in gray and excluded from CFN scoring. The same applies hereafter.}
  \label{fig:cfn_heatmap_4b_0854}
\end{figure}

\begin{figure*}[p]
  \centering
  \includegraphics[width=0.87\textwidth]{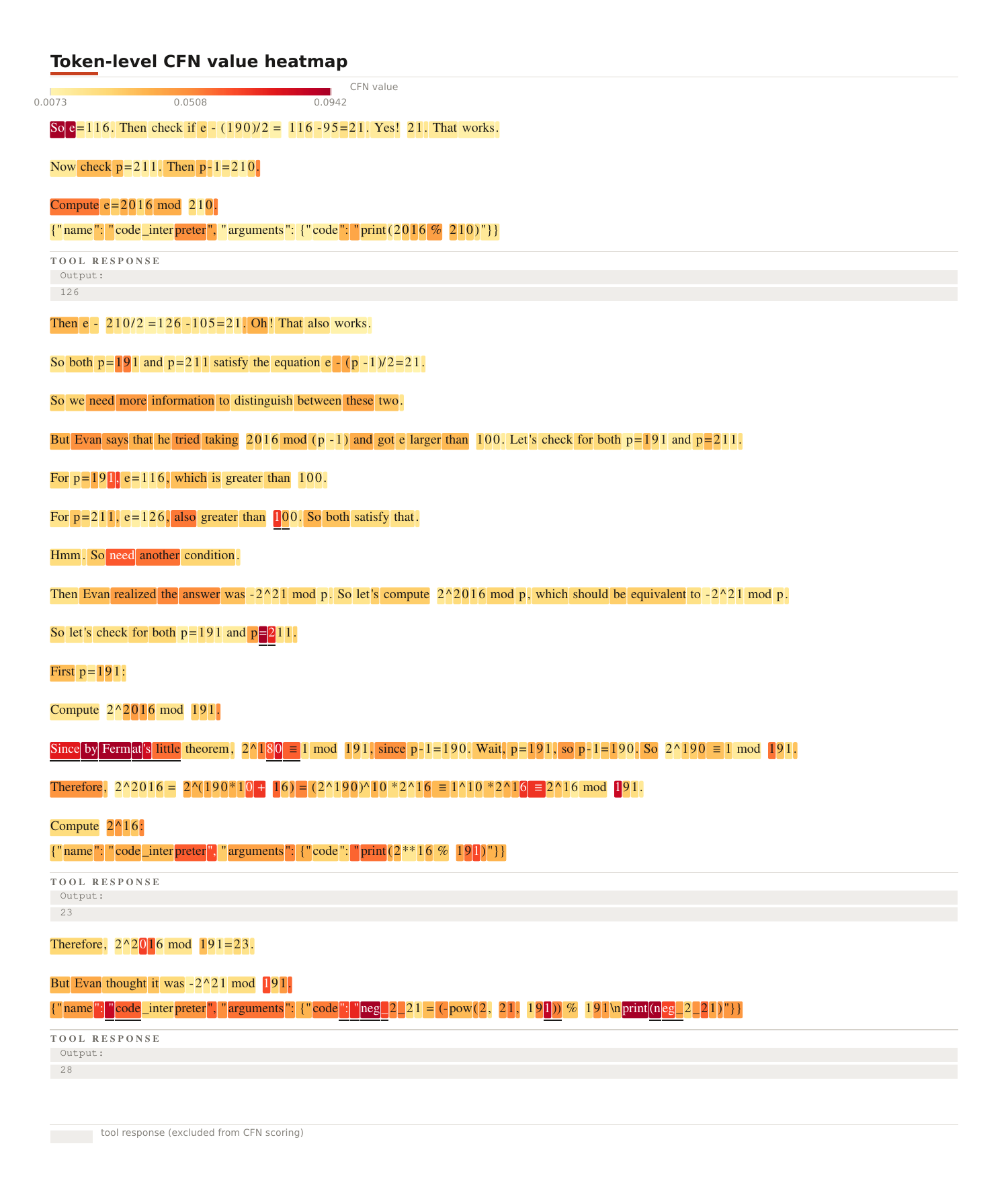}
  \caption{An example of token-level uncertainty estimated by CFN within a Qwen3-4B modular-arithmetic response.}
  \label{fig:cfn_heatmap_4b_088}
\end{figure*}

\begin{figure*}[p]
  \centering
  \includegraphics[width=0.97\textwidth]{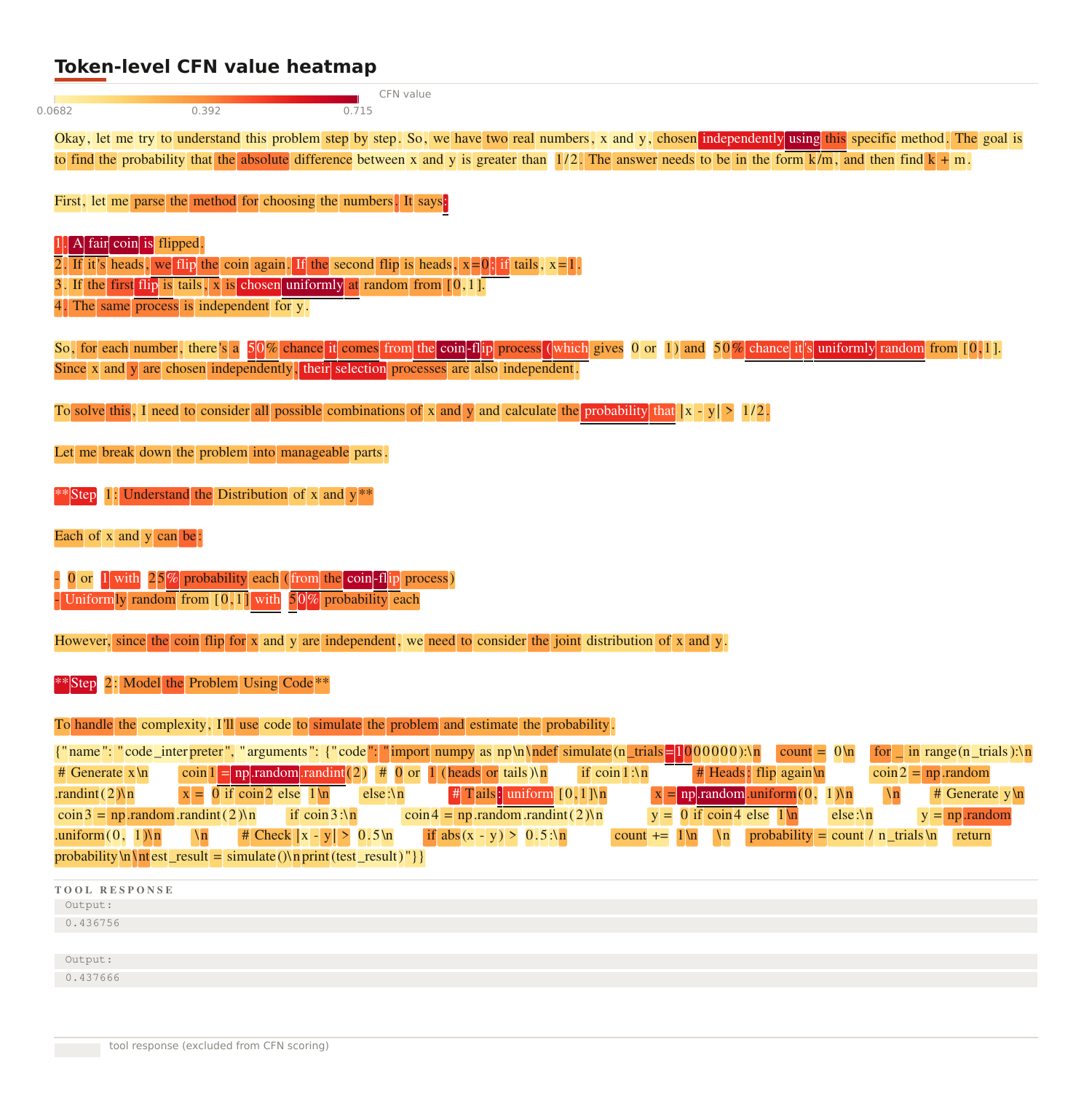}
  \caption{An example of token-level uncertainty estimated by CFN within a Qwen2.5-7B modular-arithmetic response.}
  \label{fig:cfn_heatmap_7b_03}
\end{figure*}

\begin{figure*}[p]
  \centering
  \includegraphics[width=0.87\textwidth]{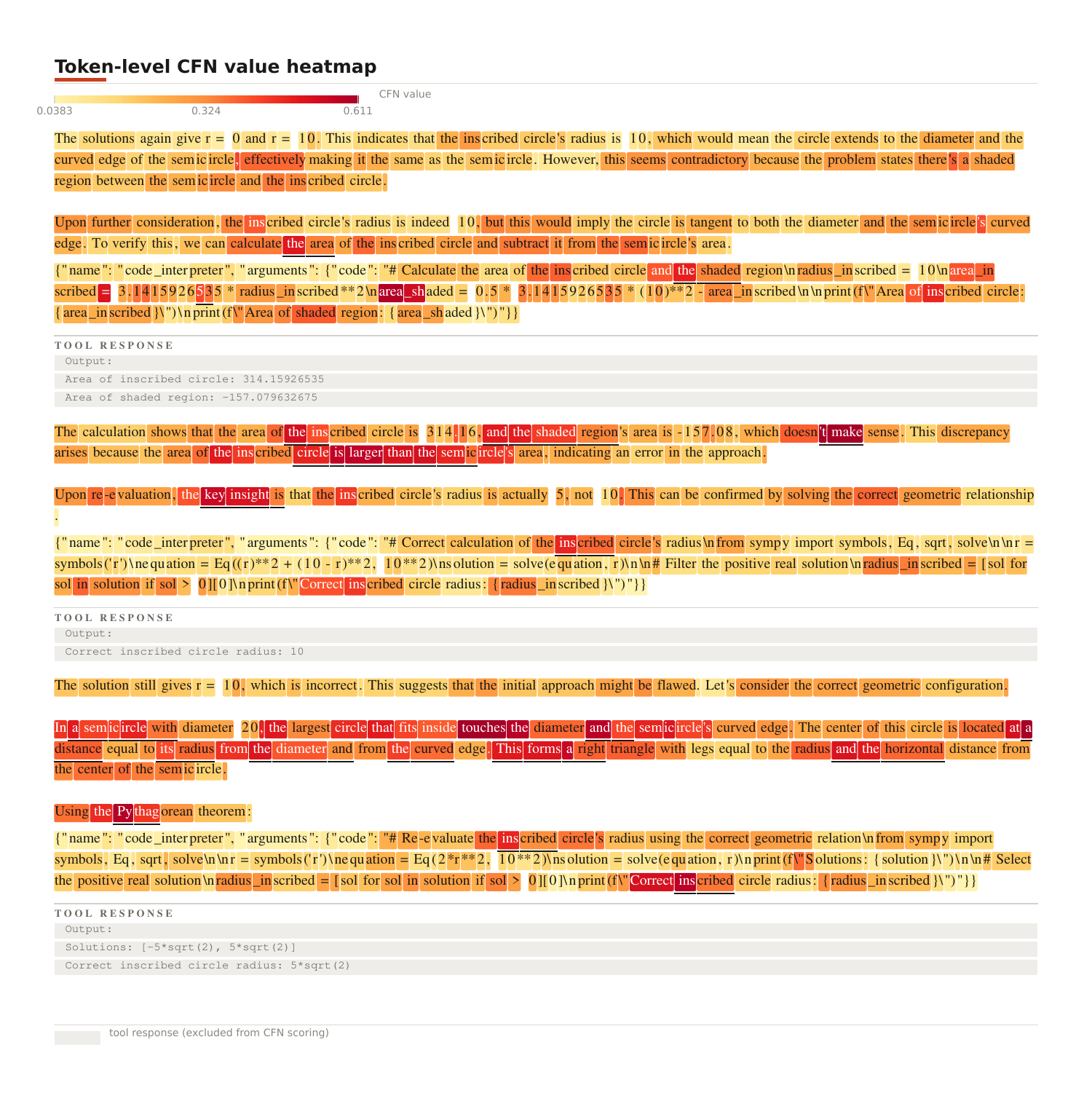}
  \caption{An example of token-level uncertainty estimated by CFN within a Qwen2.5-7B geometric-reasoning response.}
  \label{fig:cfn_heatmap_7b_037}
\end{figure*}

\begin{figure*}[p]
  \centering
  \includegraphics[width=0.87\textwidth]{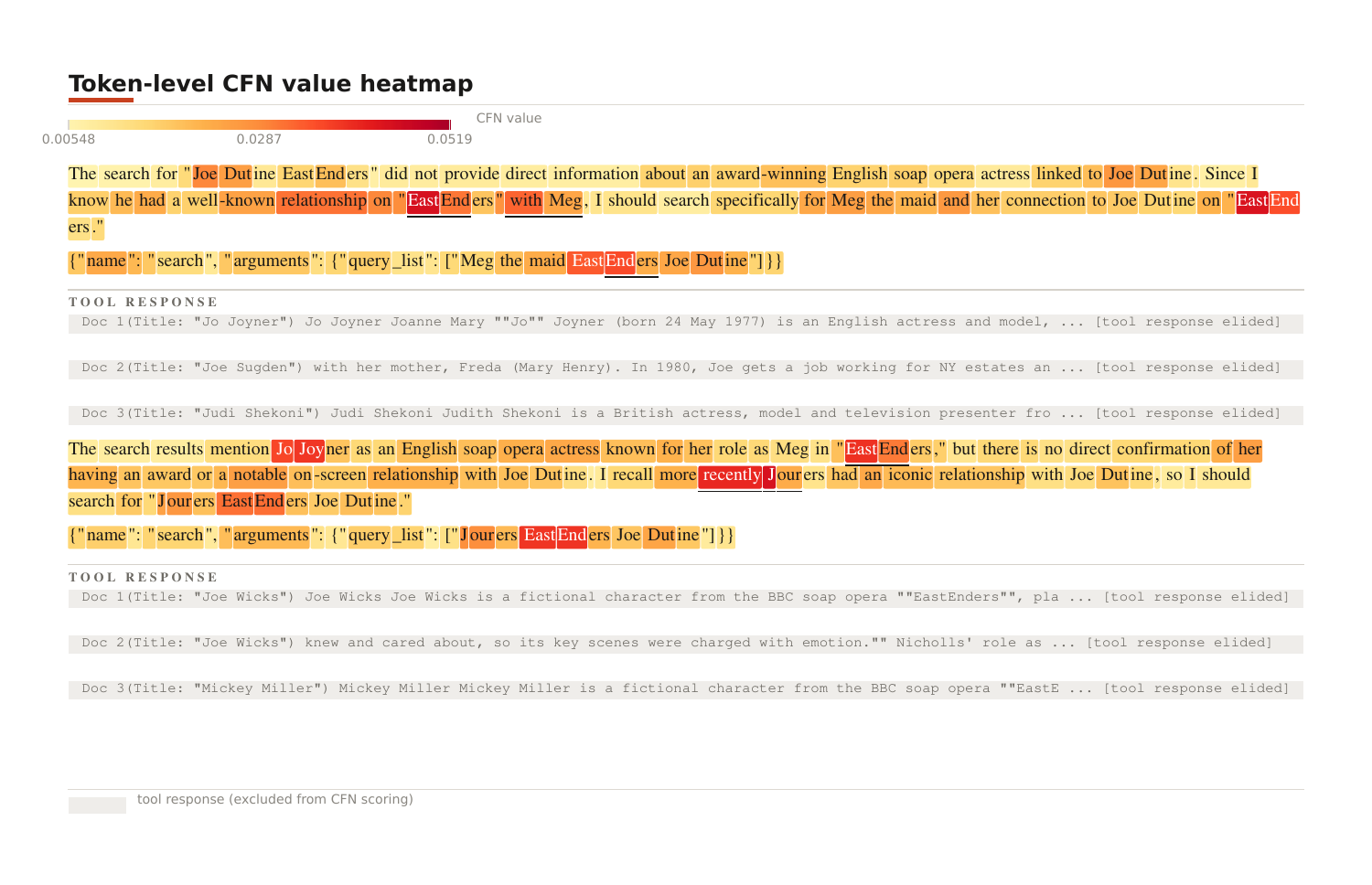}
  \caption{An example of token-level uncertainty estimated by CFN within a Qwen2.5-7B-Instruct search response.}
  \label{fig:cfn_heatmap_7b_search}
\end{figure*}

\begin{figure*}[p]
  \centering
  \includegraphics[width=0.87\textwidth]{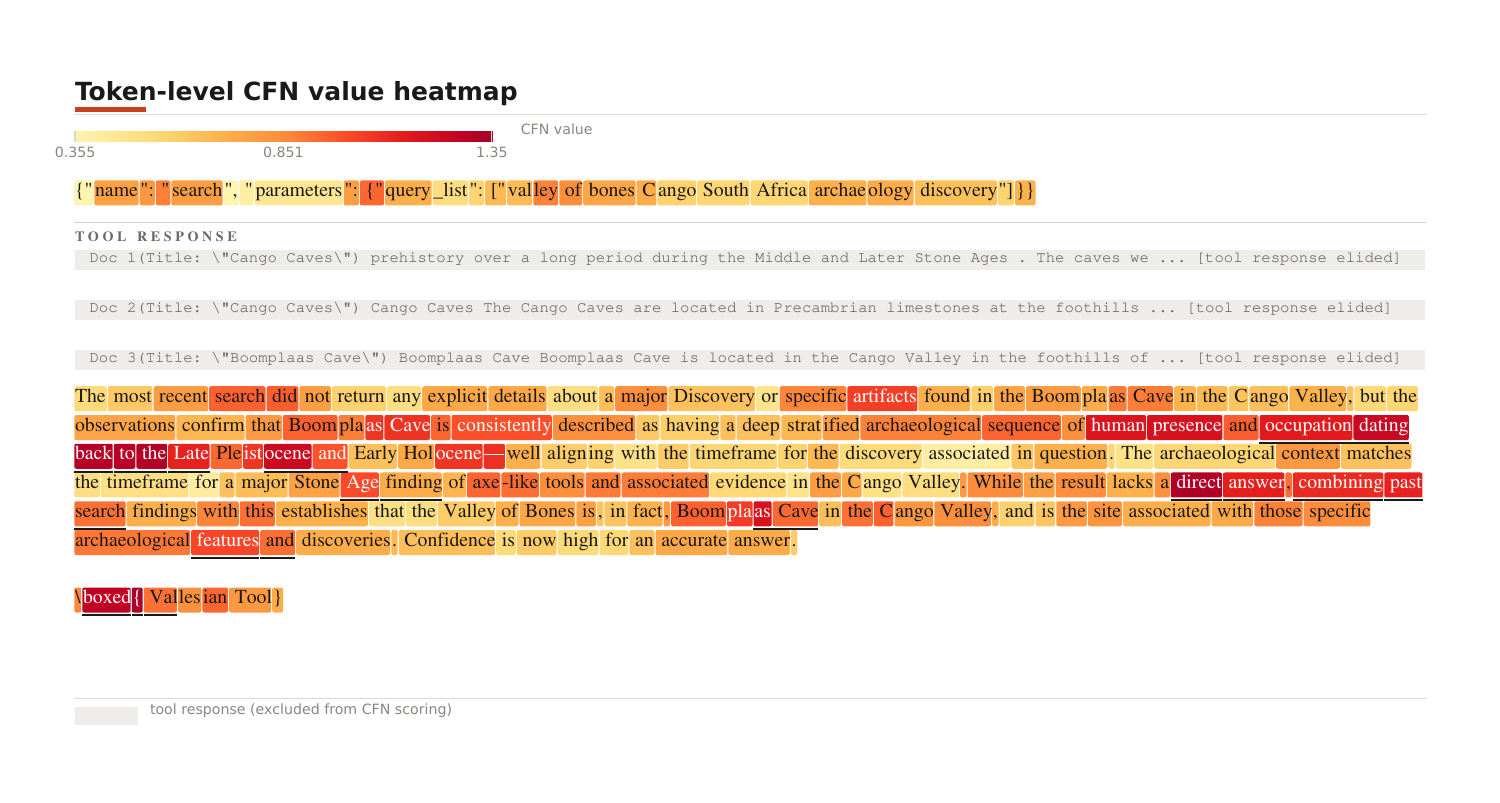}
  \caption{An example of token-level uncertainty estimated by CFN within a Llama-3.2-3B-Instruct search response.}
  \label{fig:cfn_heatmap_llama_search}
\end{figure*}
\clearpage

Across the examples, uncertainty changes within each response.
In Figure~\ref{fig:cfn_heatmap_4b_0854}, high scores appear when the model changes its plan from direct modulo calculations to a different exponentiation method, and when it builds a new tool call.
In Figure~\ref{fig:cfn_heatmap_4b_088}, high scores appear when the model compares two possible values of $p$, applies Fermat's little theorem, and later fixes its calculation.
In Figure~\ref{fig:cfn_heatmap_7b_03}, high scores appear when the model works through a sampling process that combines discrete and continuous choices and turns it into simulation code.
Figure~\ref{fig:cfn_heatmap_7b_037} shows another geometry example: high scores appear when the model notices that its first radius calculation cannot be right, rethinks the geometry, and writes the correct equation using the Pythagorean theorem.
In Figure~\ref{fig:cfn_heatmap_7b_search}, high scores concentrate on named entities and relation-bearing terms as the model reformulates its query after the initial search results do not resolve the target relation.
In Figure~\ref{fig:cfn_heatmap_llama_search}, high scores appear when the model combines retrieved archaeological evidence, reasons over its temporal context, and commits to the final answer despite the absence of a direct match.
By contrast, common connecting words, routine calculations, and repeated search scaffolding generally receive lower scores.
The high scores usually cover several neighboring tokens rather than isolated tokens.
Across both task types, these high-score spans correspond to changes in reasoning strategy, tool-query formulation, evidence synthesis, code, or specialized mathematical steps.
This suggests that the learned CFN head captures high-level reasoning patterns in both internal reasoning and tool-augmented information seeking, rather than only token-level differences.
Each heatmap has its own score range, so its colors should be read within that example rather than compared directly with the other figures.

\subsubsection{Reasoning-Pattern Sensitivity in Rollout Trees}
\label{app:cfn_semantic_sensitivity}
During model rollouts, we record each expanded parent node together with its
children and their learned pseudo-counts. We sample some instances from these
rollout records for qualitative visualization. Each instance contains two
textually different children that follow the same or a closely related
reasoning pattern, together with a third child that takes a more distinct
approach.

\begin{figure}[!htbp]
  \centering
  \includegraphics[width=0.95\linewidth]{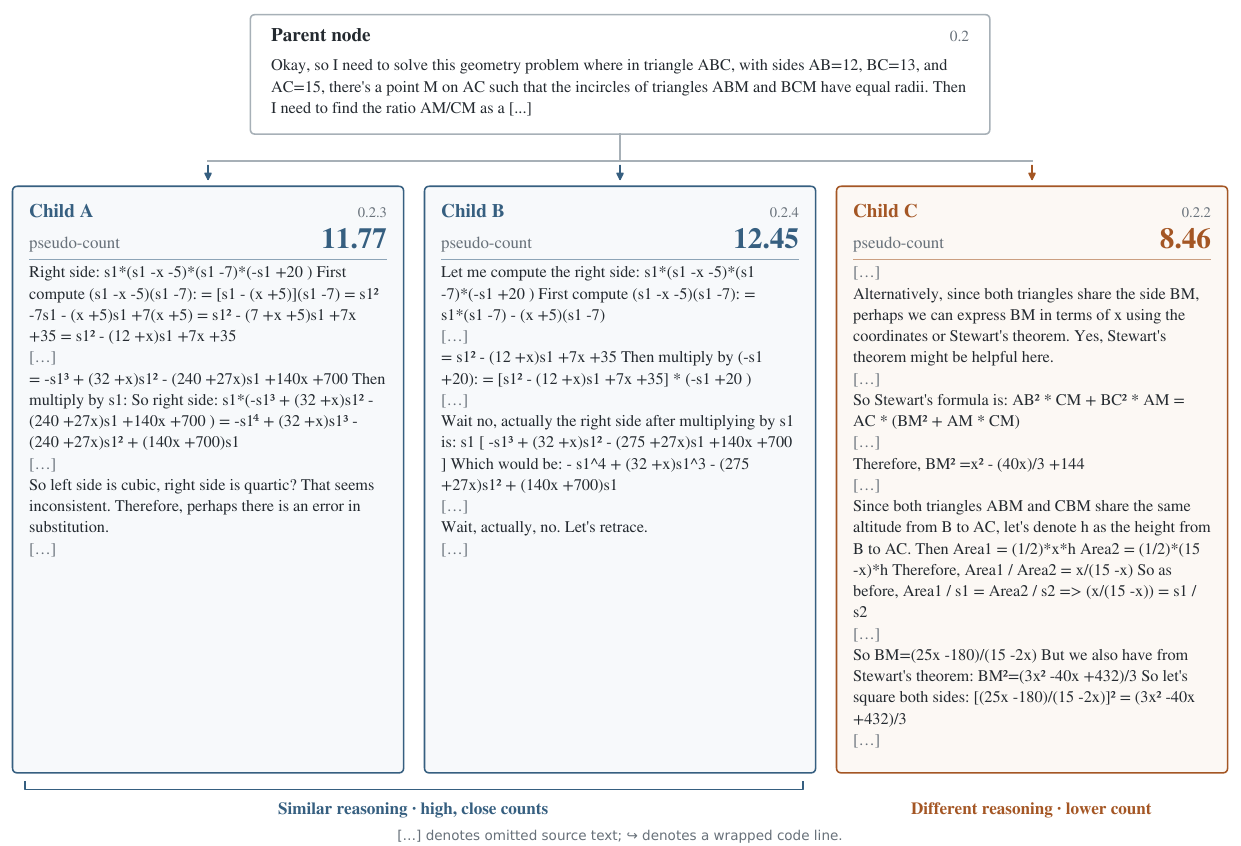}
  \caption{\textbf{A geometry case from an actual Qwen3-4B rollout tree.}
  Children A and B continue the same algebraic expansion and receive close
  pseudo-counts of 11.77 and 12.45. Child C instead combines Stewart's theorem
  with an area argument and receives a lower pseudo-count of 8.46.}
  \label{fig:cfn_rollout_case_01}
\end{figure}

\begin{figure*}[p!]
  \centering
  \includegraphics[width=0.92\textwidth]{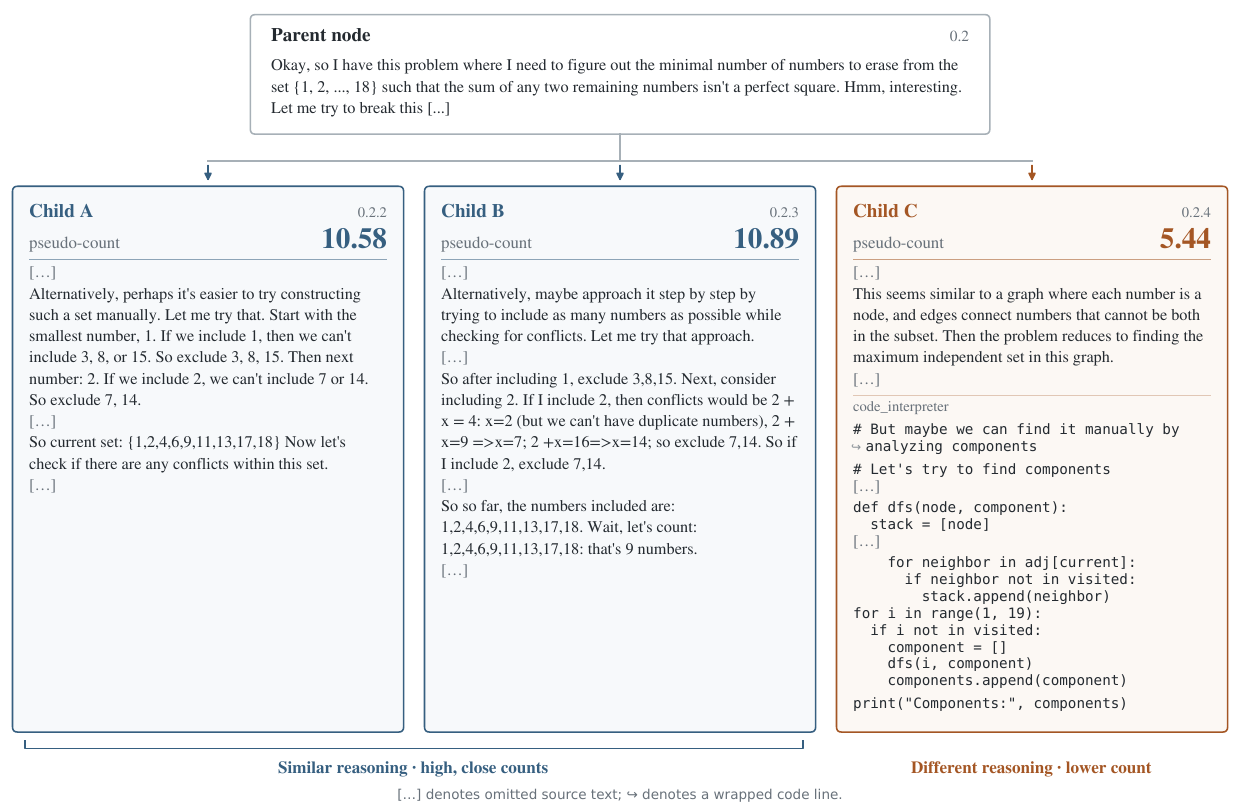}
  \caption{\textbf{A combinatorial-reasoning case from an actual Qwen3-4B
  rollout tree.} Children A and B repeat closely related manual constructions
  and receive pseudo-counts of 10.58 and 10.89. Child C reformulates the task as
  a maximum-independent-set problem and receives a lower count of 5.44.}
  \label{fig:cfn_rollout_case_03}
\end{figure*}

\begin{figure*}[p!]
  \centering
  \includegraphics[width=0.92\textwidth]{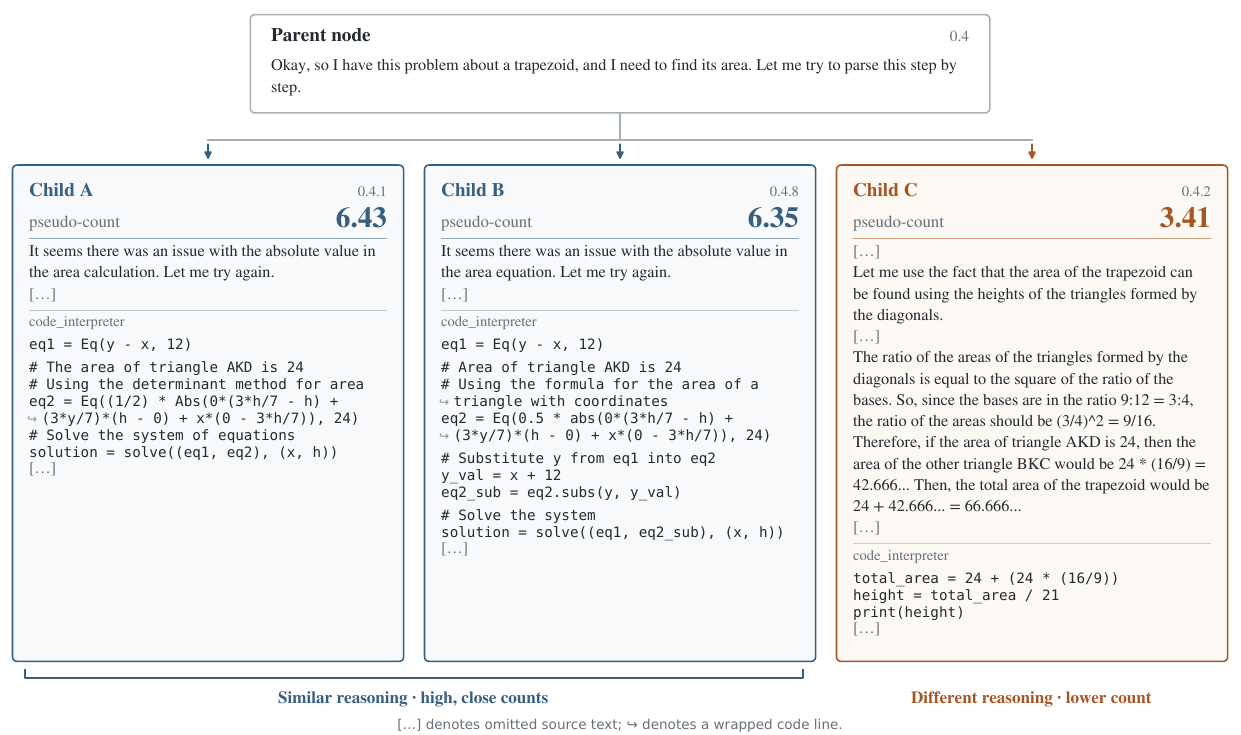}
  \caption{\textbf{A geometry case from an actual Qwen2.5-7B rollout tree.}
  Children A and B retry closely related coordinate-based calculations and
  receive pseudo-counts of 6.43 and 6.35. Child C replaces this route with an
  area-ratio argument and receives a lower count of 3.41.}
  \label{fig:cfn_rollout_case_05}
\end{figure*}

\begin{figure*}[p!]
  \centering
  \includegraphics[width=0.95\textwidth]{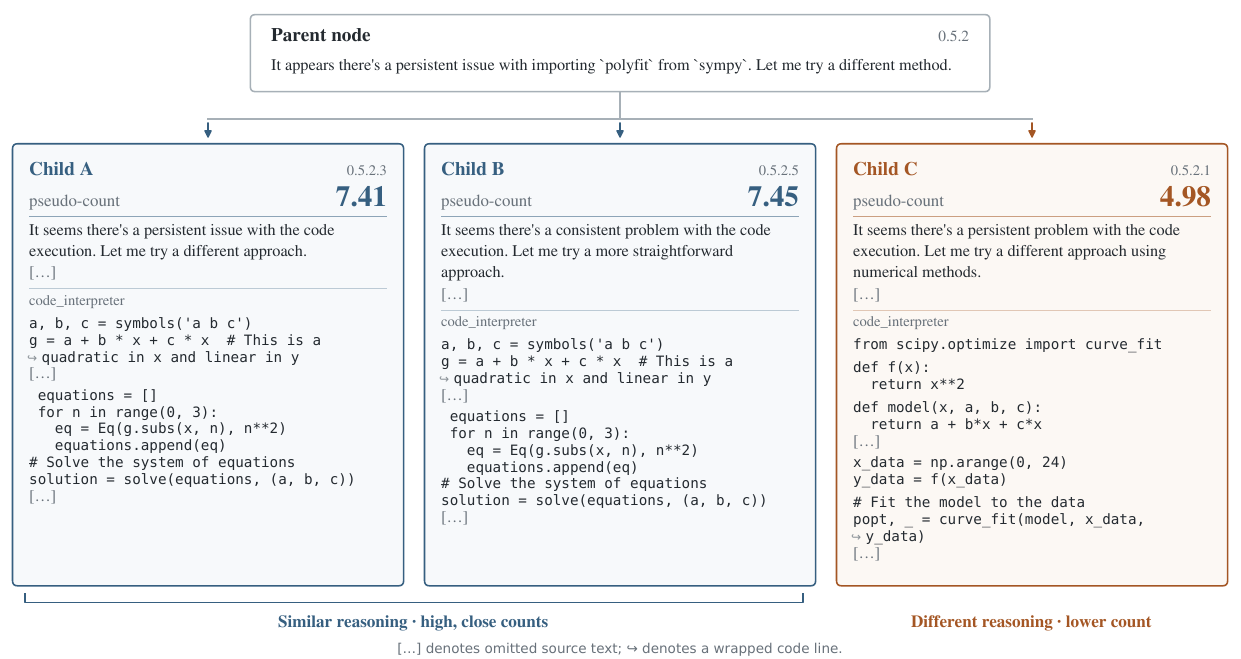}
  \caption{\textbf{A symbolic-computation case from an actual Qwen2.5-7B
  rollout tree.} Children A and B repeat the same symbolic-solving attempt and
  receive nearly identical pseudo-counts of 7.41 and 7.45. Child C switches to
  numerical fitting and receives a lower count of 4.98.}
  \label{fig:cfn_rollout_case_06}
\end{figure*}

\begin{figure*}[p!]
  \centering
  \includegraphics[width=0.95\textwidth]{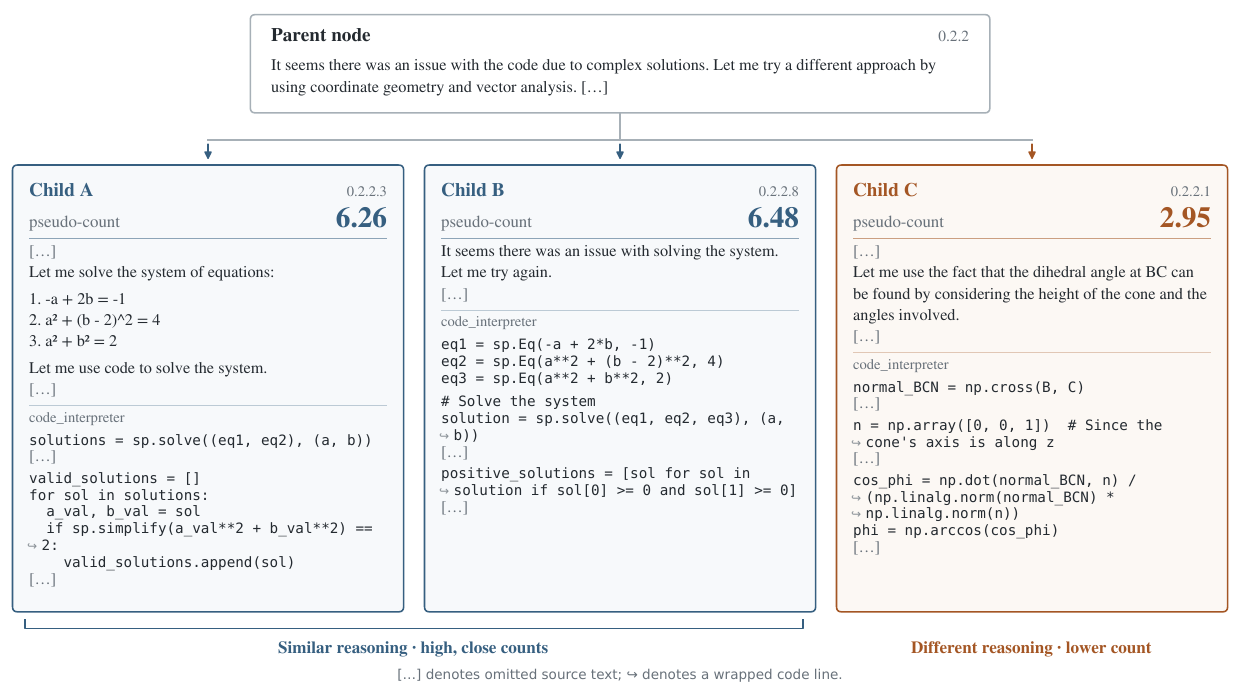}
  \caption{\textbf{A solid-geometry case from an actual Qwen2.5-7B rollout tree.}
  Children A and B retry the same coordinate-system equations and receive
  pseudo-counts of 6.26 and 6.48. Child C computes the dihedral angle with a
  vector normal and receives a lower count of 2.95.}
  \label{fig:cfn_rollout_additional_03}
\end{figure*}

\begin{figure*}[p!]
  \centering
  \includegraphics[width=0.95\textwidth]{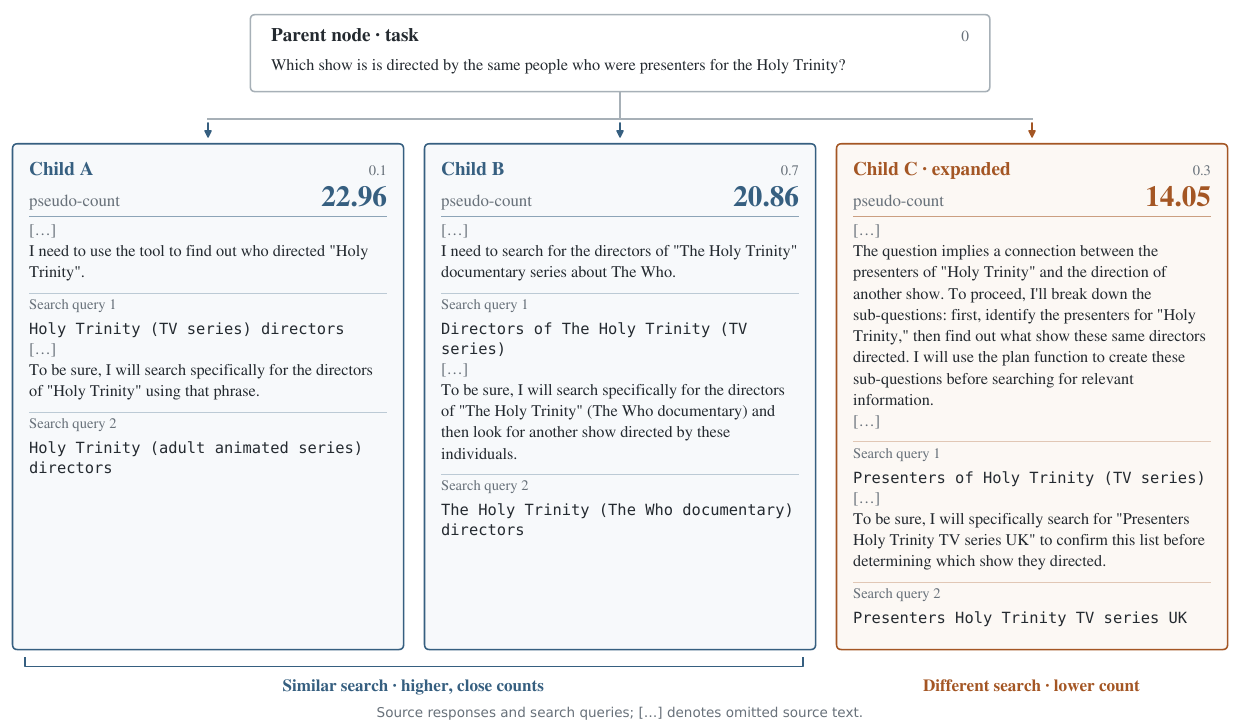}
  \caption{\textbf{A query-planning case from an actual Qwen2.5-7B-Instruct search rollout tree.}
  Children A and B repeatedly search for the program's directors and receive
  close pseudo-counts of 22.96 and 20.86. Child C instead follows the presenter
  relation first and receives a lower count of 14.05.}
  \label{fig:cfn_rollout_search_7b}
\end{figure*}

\begin{figure}[!htbp]
  \centering
  \includegraphics[width=0.95\textwidth]{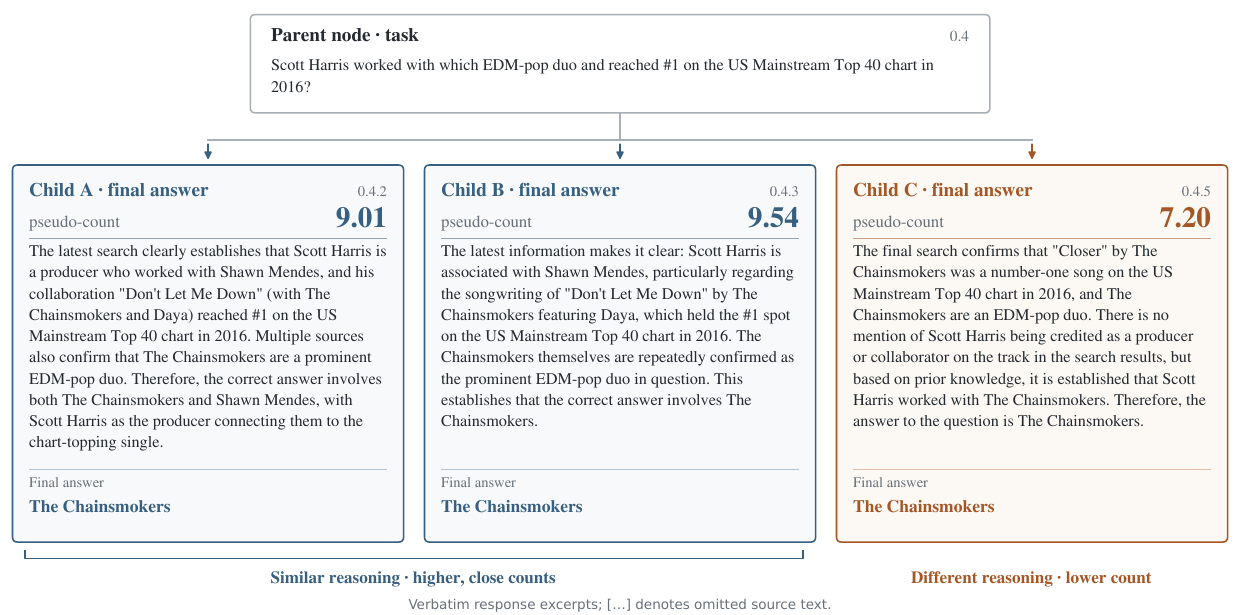}
  \caption{\textbf{An evidence-use case from an actual Llama-3.2-3B-Instruct search rollout tree.}
  Children A and B organize the same final answer around retrieved relationship
  evidence and receive pseudo-counts of 9.01 and 9.54. Child C relies on prior
  knowledge after noting an evidence gap and receives a lower count of 7.20.}
  \label{fig:cfn_rollout_search_llama}
\end{figure}
% \FloatBarrier

\begin{table}[t]
  \centering
  \small
  \caption{\textbf{Semantic alignment of learned pseudo-counts.} Spearman correlations are reported with 95\% confidence intervals (CIs). Nearest-neighbor agreement compares count closeness with embedding cosine similarity.}
  \label{tab:cfn_semantic_alignment}
  \resizebox{0.7\linewidth}{!}{%
  \begin{tabular}{llcr}
    \toprule
    \Mod{CFN backbone} & \Mod{Embedding model} & \Mod{Spearman $\rho$ [95\% CI]} & \Mod{\shortstack{Nearest-neighbor\\agreement}} \\
    \midrule
    Qwen3-4B & Qwen3-Embedding-4B & 0.255 [0.197, 0.315] & 24.1\% \\
    Qwen3-4B & QZhou-Embedding & 0.248 [0.197, 0.299] & 24.8\% \\
    \midrule
    Qwen2.5-7B & Qwen3-Embedding-4B & 0.335 [0.279, 0.391] & 29.2\% \\
    Qwen2.5-7B & QZhou-Embedding & 0.358 [0.303, 0.413] & 26.9\% \\
    \bottomrule
  \end{tabular}%
  }
\end{table}

Together, Figure~\ref{fig:tree_local_vs_learned} and
Figures~\ref{fig:cfn_rollout_case_01}--\ref{fig:cfn_rollout_search_llama}
show this behavior across eight sampled rollout-tree examples.
The six mathematical-reasoning cases cover changes in theorem, representation, and computational strategy, while the two search cases cover query planning and evidence use.
Across both tasks and model families, textually different children following the same high-level pattern receive close pseudo-counts, whereas a child that changes the reasoning pattern shows a substantially larger pseudo-count gap from the aligned pair.
The final search case further shows that this distinction can persist even when all children produce the same answer, indicating that the estimator responds to the reasoning path rather than outcome identity alone.

This behavior addresses a limitation of standard MCTS counts. Standard MCTS
maintains an independent tabular count for each child, so sampling child A does
not increase the count of a semantically related child B. Consequently, B can
still appear unexplored even when its reasoning pattern has already been
covered through A. LPC-MCTS instead assigns pseudo-counts from learned hidden
representations, allowing such branches to share familiarity. Holding the
other UCT terms fixed, the larger counts of A and B reduce redundant exploration
of the repeated pattern, while the lower count of C gives the distinct strategy
a larger exploration bonus.

\subsubsection{\Mod{Quantitative Mechanism Validation}}
\label{app:cfn_mechanism_validation}
Both experiments use frozen Qwen3-4B and Qwen2.5-7B backbones on mathematical reasoning tasks.

\paragraph{\Mod{Semantic alignment.}}
For each backbone, we sample 8 distinct non-root MCTS nodes from each of 128 questions, independently of pseudo-counts or embedding similarities.
For node pairs within each question, we compare count closeness, $-|\log C_\phi(v_i)-\log C_\phi(v_j)|$, with cosine similarity between embeddings of the generated text.
Using Qwen3-Embedding-4B and QZhou-Embedding, we report Spearman correlation and nearest-neighbor agreement.

\autoref{tab:cfn_semantic_alignment} shows positive correlations ($\rho=0.248$--$0.358$; all 95\% CIs exclude zero) and 24.1--29.2\% nearest-neighbor agreement, above the 14.3\% uniform-random baseline.
These results support the semantic relevance of learned pseudo-counts.

\paragraph{\Mod{Count transfer across branches.}}
For sibling responses $A$ and $B$, we use GPT-6 to label reasoning-pattern similarity separately from content equivalence.
We measure the change in the pseudo-count of held-out sibling $B$ after 16 coin-flip updates to the CFN head on $A$.
Each trial starts from the original head and a fresh optimizer.
Log-count changes are averaged within questions and reported as geometric mean percentage changes with question-level bootstrap 95\% CIs.

\begin{table}[t]
  \centering
  \small
  \caption{\textbf{Count transfer to held-out sibling responses.} Results across Qwen3-4B and Qwen2.5-7B measure the change in a probe's pseudo-count after CFN updates on its sibling.}
  \label{tab:cfn_count_transfer}
  \begin{tabular}{lrc}
    \toprule
    \Mod{Probe nodes} & \Mod{Questions} & \Mod{\shortstack{Pseudo-count change\\(geometric mean; 95\% CI)}} \\
    \midrule
    Same-pattern siblings & 400 & $+5.5\%\;[+3.5\%,+7.4\%]$ \\
    Distinct-pattern siblings & 126 & $-3.2\%\;[-5.0\%,-1.3\%]$ \\
    \bottomrule
  \end{tabular}
\end{table}

\autoref{tab:cfn_count_transfer} shows a 5.5\% increase for same-pattern siblings and a 3.2\% decrease for distinct-pattern siblings, with both 95\% CIs excluding zero.
This supports selective evidence sharing across related reasoning branches.

\FloatBarrier

\begin{figure*}[t]
  \centering
  \includegraphics[width=\textwidth]{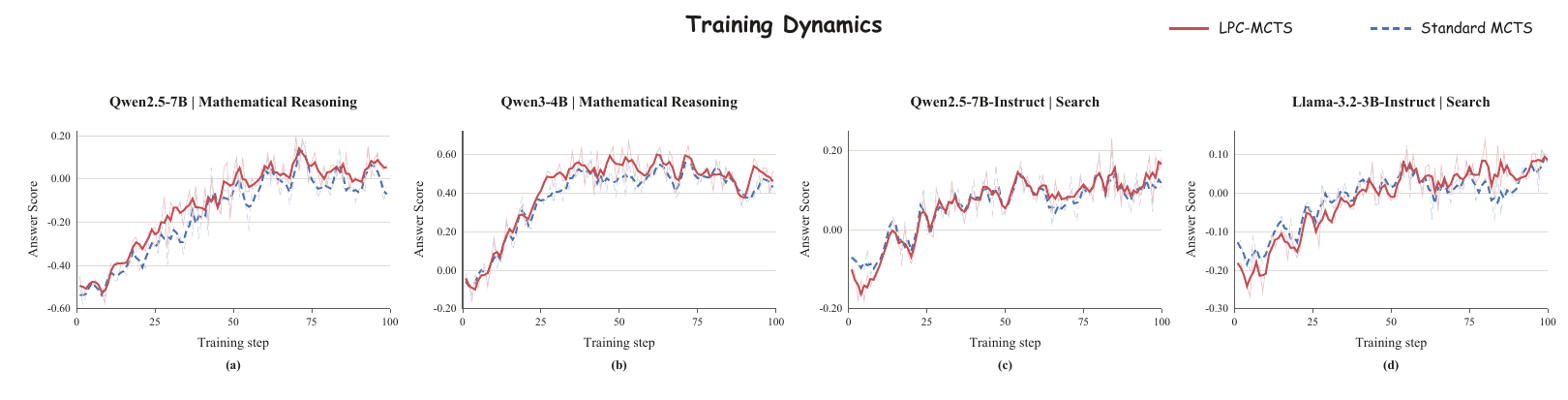}
  \caption{Training reward dynamics of standard MCTS and LPC-MCTS over 100
  policy-training steps across two mathematical-reasoning and two search
  model settings.}
  \label{fig:training_dynamics}
\end{figure*}

\subsection{MCTS Training}
\subsubsection{Training Dynamics}
Figure~\ref{fig:training_dynamics} compares the answer scores of standard
MCTS and LPC-MCTS across four task--model settings at training time.
The curves are smoothed using a 3-step running average.

\subsubsection{Ablation Studies}
\label{app:ablation}
To separate the effects of the training and inference configurations, Table~\ref{tab:math_ablation} specifies the method used at each stage.
Pseudo-count-guided trajectory selection is used only to construct the policy-training batch, so we ablate it only during training by removing the ranking and selection rule in Eq.~\ref{eq:trajectory_selection}.
Token-to-node aggregation is used to compute node pseudo-counts in both stages and is therefore ablated separately during training and inference.
When aggregation is removed at a given stage, let $t_{\mathrm{last}}$ be the last eligible model-generated token in node $v$.
We directly use its CFN score to compute the node-level pseudo-count:
\begin{equation}
    U_\phi(v)=u_{v,t_{\mathrm{last}}},
    \qquad
    C_\phi(v)=\frac{1}{U_\phi(v)+\epsilon}.
\end{equation}
The combined training variant removes both trajectory selection and node aggregation.
We also remove node aggregation at both stages to assess its end-to-end effect.
The w/o pseudo-count variant uses standard visit counts for tree search during training and inference, while retaining pseudo-count-guided trajectory selection for policy training.
All other settings are kept identical to full LPC-MCTS.

\begin{table*}[t]
  \caption{Ablation results under separate training and inference settings on tool-integrated mathematical reasoning with Qwen2.5-7B and Qwen3-4B.
  Each value is the mean accuracy over 16 responses per problem (mean@16).}
  \label{tab:math_ablation}
  \begin{center}
  \resizebox{\textwidth}{!}{%
  \begin{tabular}{llcccccc}
    \toprule
    Training & Inference & AIME 24 & AIME 25 & AIME 26 & HMMT25 & BRUMO25 & Avg. \\
    \midrule
    \multicolumn{8}{l}{\textit{Qwen2.5-7B}} \\
    \addlinespace
    \Mod{LPC-MCTS (w/o pseudo-count)} & \Mod{MCTS} & \Mod{39.17} & \Mod{25.63} & \Mod{24.79} & \Mod{18.96} & \Mod{44.58} & \Mod{30.63} \\
    LPC-MCTS (w/o trajectory selection) & LPC-MCTS & 34.79 & 29.58 & 27.71 & 15.83 & 42.50 & 30.08 \\
    LPC-MCTS (w/o node aggregation) & LPC-MCTS & 37.71 & 28.75 & 25.63 & 19.17 & 48.54 & 31.96 \\
    LPC-MCTS (w/o both) & LPC-MCTS & 37.50 & 26.04 & 28.96 & 18.54 & 45.42 & 31.29 \\
    LPC-MCTS & LPC-MCTS (w/o node aggregation) & 35.83 & 28.33 & 27.08 & 19.17 & 46.67 & 31.42 \\
    LPC-MCTS (w/o node aggregation) & LPC-MCTS (w/o node aggregation) & 38.13 & 27.08 & 27.08 & 18.13 & 41.88 & 30.46 \\
    LPC-MCTS & LPC-MCTS & 37.08 & 30.63 & 27.92 & 23.33 & 45.00 & 32.79 \\
    \midrule[\heavyrulewidth]
    \multicolumn{8}{l}{\textit{Qwen3-4B}} \\
    \addlinespace
    \Mod{LPC-MCTS (w/o pseudo-count)} & \Mod{MCTS} & \Mod{76.25} & \Mod{57.71} & \Mod{64.38} & \Mod{47.29} & \Mod{65.21} & \Mod{62.17} \\
    LPC-MCTS (w/o trajectory selection) & LPC-MCTS & 74.38 & 63.13 & 72.08 & 50.21 & 65.63 & 65.09 \\
    LPC-MCTS (w/o node aggregation) & LPC-MCTS & 75.21 & 62.71 & 70.83 & 48.33 & 62.08 & 63.83 \\
    LPC-MCTS (w/o both) & LPC-MCTS & 70.63 & 58.13 & 66.04 & 47.29 & 64.79 & 61.38 \\
    LPC-MCTS & LPC-MCTS (w/o node aggregation) & 74.79 & 62.50 & 70.42 & 48.96 & 65.83 & 64.50 \\
    LPC-MCTS (w/o node aggregation) & LPC-MCTS (w/o node aggregation) & 73.13 & 58.75 & 67.29 & 49.38 & 62.50 & 62.21 \\
    LPC-MCTS & LPC-MCTS & 78.13 & 63.54 & 72.92 & 51.88 & 66.04 & 66.50 \\
    \bottomrule
  \end{tabular}%
  }
  \end{center}
\end{table*}

Table~\ref{tab:math_ablation} shows that the ablated components contribute across backbones.
Removing pseudo-counts from tree search while retaining trajectory selection lowers the average by 2.16 points on Qwen2.5-7B and 4.33 points on Qwen3-4B.
Removing pseudo-count-guided trajectory selection lowers the average
by 2.71 points on Qwen2.5-7B and 1.41 points on Qwen3-4B. Removing
token-to-node aggregation during training lowers the average by 0.83 and 2.67
points, while removing it only at inference lowers it by 1.37 and 2.00 points.
Removing aggregation at both stages produces larger drops of 2.33 and 4.29
points, respectively. Removing both training components also reduces the
average by 1.50 and 5.12 points. These consistent average degradations support
the roles of trajectory selection in policy training and node aggregation in
both training and inference.

\subsubsection{Hyperparameters}
\label{app:filtering_hyperparameters}
The default token-filtering ratio in LPC-MCTS is $p=30$, as described in Appendix~\ref{app:token_aggregation}.
To examine sensitivity to this choice, we additionally evaluate $p=50$, $p=70$, and $p=100$.
A larger $p$ retains a broader set of token-level CFN scores, while $p=100$ retains all eligible tokens before contiguous-run filtering.
The same value of $p$ is used during training and inference, with all other configurations fixed.

\begin{table}[!htbp]
  \caption{Sensitivity to the top-$p\%$ token-filtering ratio on tool-integrated mathematical reasoning with Qwen2.5-7B and Qwen3-4B.
  Each value is the mean accuracy over 16 responses per problem (mean@16).}
  \label{tab:filtering_hyperparameters}
  \begin{center}
  \resizebox{0.7\linewidth}{!}{%
  \begin{tabular}{lcccccc}
    \toprule
    Top $p\%$ & AIME 24 & AIME 25 & AIME 26 & HMMT25 & BRUMO25 & Avg. \\
    \midrule
    \multicolumn{7}{l}{\textit{Qwen2.5-7B}} \\
    \addlinespace
    $30\%$ & 37.08 & 30.63 & 27.92 & 23.33 & 45.00 & 32.79 \\
    $50\%$ & 36.25 & 26.67 & 28.75 & 22.71 & 45.00 & 31.88 \\
    $70\%$ & 34.38 & 31.67 & 27.08 & 24.79 & 42.71 & 32.13 \\
    $100\%$ & 38.96 & 29.17 & 29.79 & 22.08 & 43.33 & 32.67 \\
    \midrule
    \multicolumn{7}{l}{\textit{Qwen3-4B}} \\
    \addlinespace
    $30\%$ & 78.13 & 63.54 & 72.92 & 51.88 & 66.04 & 66.50 \\
    $50\%$ & 74.17 & 61.46 & 69.58 & 54.38 & 63.13 & 64.54 \\
    $70\%$ & 76.88 & 59.58 & 70.00 & 48.96 & 66.25 & 64.33 \\
    $100\%$ & 78.96 & 60.21 & 71.25 & 48.54 & 62.29 & 64.25 \\
    \bottomrule
  \end{tabular}%
  }
  \end{center}
\end{table}

Table~\ref{tab:filtering_hyperparameters} shows that the default $p=30$
achieves the highest average performance on both backbones. Qwen3-4B drops by
1.96--2.25 points with larger filtering ratios, whereas Qwen2.5-7B varies by
less than one point across all settings. These results support $p=30$ as a
strong default while indicating that LPC-MCTS is not highly sensitive to this
choice, especially on Qwen2.5-7B. Moreover, retaining all eligible tokens
($p=100$) does not improve the average, suggesting that filtering helps exclude
less informative token-level estimates from node aggregation.

\subsection{\Mod{Inference-Time Cost Analysis}}
\label{app:inference_cost}

We compare MCTS and LPC-MCTS on the 60 problems from AIME 2024 and AIME 2025 using 2 GPUs and 16 candidate slots per problem.
For each backbone, both methods use the same policy checkpoint, decoding parameters, and maximum search-iteration budget; the CFN head is frozen during inference.

\autoref{tab:inference_cost} reports the overall inference costs.
Policy-generated token counts cover the entire search, including branches not selected for the final output. The amortized time per candidate includes all search and CFN computation, but excludes model initialization and post-processing.

\begin{table}[t]
  \centering
  \small
  \setlength{\tabcolsep}{4pt}
  \caption{\textbf{Inference-time costs on AIME 2024/2025.} Generated tokens are summed over the entire search; time is the amortized wall-clock cost per candidate slot.}
  \label{tab:inference_cost}
  \begin{tabular}{lrrrrr}
    \toprule
    & \multicolumn{2}{c}{\Mod{Generated tokens (M)}} & & \multicolumn{2}{c}{\Mod{Time (s/candidate)}} \\
    \cmidrule(lr){2-3}\cmidrule(lr){5-6}
    \Mod{Model} & \Mod{MCTS} & \Mod{LPC-MCTS} & \Mod{Reduction} & \Mod{MCTS} & \Mod{LPC-MCTS} \\
    \midrule
    Qwen3-4B & 80.958 & 63.106 & 22.05\% & 16.72 & 15.98 \\
    Qwen2.5-7B & 44.007 & 25.525 & 42.00\% & 12.05 & 12.83 \\
    \bottomrule
  \end{tabular}
\end{table}

LPC-MCTS generates 22.05\% fewer tokens on Qwen3-4B and 42.00\% fewer on Qwen2.5-7B under the same search budget, showing the higher token efficiency.
The amortized time per candidate falls from 16.72 to 15.98 seconds on Qwen3-4B, and rises from 12.05 to 12.83 seconds on Qwen2.5-7B.
Thus, the measured wall-clock cost remains similar in these runs, despite introducing a CFN head.

\autoref{tab:cfn_inference_cost} shows where the CFN spends time.
For each component, we sum its elapsed time across workers and divide by the total search time across those workers.
Feature extraction includes requesting hidden states from the backbone, waiting for the request, and processing the returned states; scoring/control includes evaluating the CFN head.

\begin{table}[t]
  \centering
  \small
  \caption{\textbf{CFN workload and runtime breakdown.} Time shares use summed worker search time as the denominator for LPC-MCTS under the same evaluation setting as \autoref{tab:inference_cost}.}
  \label{tab:cfn_inference_cost}
  \begin{tabular}{lrr}
    \toprule
    \Mod{Measurement} & \Mod{Qwen3-4B} & \Mod{Qwen2.5-7B} \\
    \midrule
    Mean CFN input length (tokens/sequence) & 2,905.48 & 1,442.85 \\
    CFN input sequences processed & 20,968 & 16,512 \\
    CFN feature-extraction time share & 6.50\% & 5.31\% \\
    CFN scoring/control time share & 0.23\% & 0.18\% \\
    Combined CFN time share & 6.73\% & 5.49\% \\
    \bottomrule
  \end{tabular}
\end{table}

CFN-related work accounts for 6.73\% of total worker search time on Qwen3-4B and 5.49\% on Qwen2.5-7B.
Most of this time is spent obtaining hidden states from the backbone, compared with just 0.23\% and 0.18\% for CFN scoring/control.

\end{document}